\documentclass[lettersize,journal]{IEEEtran}
\usepackage{subcaption}
\usepackage{multirow} 
\usepackage{comment}
\usepackage{graphicx}
\usepackage{hyperref}
\usepackage{balance}
\usepackage{makecell}

\usepackage[noadjust]{cite}

\begin{document}
%\title{Parametric Enhancement of PerceptNet: A Human-Inspired Approach for Image Quality Assessment}

%\title{Parametric PerceptNet: A tractable bioinspired deep-net \\ for Image Quality Assessment}

\title{Parametric PerceptNet: A bioinspired deep-net \\ trained for Image Quality Assessment}

\author{Jorge Vila-Tomás\textsuperscript{1}, Pablo Hernández-Cámara\textsuperscript{1}, Valero Laparra\textsuperscript{1}, Jesús Malo\textsuperscript{1} \\
\textit{\textsuperscript{1}Image Processing Lab, Universitat de València, Paterna, 46980, Spain}}

\maketitle

\begin{abstract}
Human vision models are at the core of image processing. For instance, classical approaches to the problem of image quality are based on models that include knowledge about human vision. However, nowadays, deep learning approaches have obtained competitive results by simply approaching this problem as regression of human decisions, and training an standard network on human-rated datasets. Deep-learning approaches have the advantage of being easily adaptable to a particular problem and they fit very efficiently when data is available. Nevertheless, mainly due to the excess of parameters, they have the problems of lacking interpretability and over-fitting.  

Here we propose a deep vision model that combines the best of both worlds by using a parametric neural network architecture. We parameterize the layers to have bioplausible functionality, and provide a set of bioplausible parameters. 

We analyzed different versions of the model and compared them with the equivalent non-parametric version. The parametric models achieve three orders of magnitude reduction in the number of parameters without suffering in regression performance. Furthermore, we show that the parametric models behave better during training and are easier to interpret from the vision science perspective. Interestingly, we find that, even initialized with bioplausible parameters, the models diverge from bioplausible solutions when trained for regression using human-rated datasets, which we call the \emph{feature-spreading problem}. 
This observation suggests that the conventional deep learning approach may be inherently flawed, and emphasizes the need to evaluate and train vision models beyond regression. 

\end{abstract}


\section{Introduction}
\label{sec:intro}

Traditionally, image processing has been based on vision science to some extent. Either directly using vision models or relying on knowledge about human perception. But also indirectly, by using scene statistics that also shaped human vision, or even practical approaches based on heuristic use of human behavior. In particular, the problem of describing the quality of an image lends itself well to be approached from the point of view of vision science, since the quality of an image is usually assessed by a human observer. For instance, the human Contrast Sensitivity Functions (CSFs)~\cite{Campbell68,Mullen85} (which are equivalent to the application of layers of center-surround neurons~\cite{DeAngelis97,Shapley11,Li92,Li22}) were proposed to be considered for image quality assessment in the 70's~\cite{Sakrison74} shortly after the definition of the CSFs in the 60's. Opponent chromatic channels~\cite{Jameson57,Shapley11,Vila23} together with Gabor-like filter-banks~\cite{Hubel59,Hubel61,Blakemore69,Simoncelli92,Olshausen96} and simple masking nonlinearities were proposed to be used together in measuring image distortions~\cite{Watson90}. These simplified (point-wise) masking nonlinearities of \emph{visual neuroscience}~\cite{NakaRushton,Legge80,Legge81} are similar to the sigmoid activation functions of the \emph{artificial neural networks} (ANNs) community~\cite{Haykin09,Goodfellow16}. Variants of these simplified (fixed) nonlineaties dominated the image quality scene in the 80's and early 90's~\cite{Barten90,Ahumada93,Daly90,Watson93libro}. However, in \emph{vision science} as opposed to conventional activation sigmoids in ANNs, these nonlinearities are known to be adaptive instead of fixed~\cite{Blakemore69,Ross91,Foley94,Foley99,Morgan06}. 
This adaptive behavior is also rooted in classical models of \emph{visual neuroscience}, either recurrent~\cite{Wilson73,Grossberg73,Amari77} or feed-forward such as the Divisive Normalization (DN)~\cite{Heeger92,Carandini94,Watson97,Carandini12,Martinez18}. These recurrent or normalization models (which have been shown to be equivalent~\cite{Grossberg78,Malo24}) were successfully included in image/video quality too through the DN~\cite{Teo94a,Malo97b,Pons99,Watson02,Laparra10,Laparra16a,Laparra17}. 

On the other hand, the Barlow hypothesis that tries to explain the organization of the visual system based on \emph{unsupervised learning}~\cite{Barlow59,Barlow01}, fueled approaches for subjective quality based on image statistics in the first decade of this century. For instance, visual similarity was related to the preservation of image statistics in the spatial domain~\cite{Wang04} or in transform domains~\cite{Moorthy11gral,Saad12} and following~\cite{schwartz_natural_2001}, metrics based on DN~\cite{Laparra10,Laparra16a} were interpreted to be implementing density factorization by capturing the structure of natural images in wavelet domains~\cite{Malo10,Malo20}. Also in a statistical vein, visual fidelity was related to information flow in noisy-wavelet models of early vision~\cite{Sheikh05,Sheikh06a}, and advances can be done by improving the information  estimates~\cite{Malo20,Malo21,Gomez19,Malo25}.

For more than 40 years, the above (interpretable models with a small number of parameters) were hand-crafted using a lot of prior biological and statistical knowledge. Interestingly, this paradigm started to change with the advent of easy-to-use automatic differentiation frameworks in the 2010's. One of the first examples of this new trend was the metric proposed in~\cite{Laparra17}. This metric was still attached to a biologically sensible and simple linear+DN model, but was end-to-end optimized to maximize the correlation with an image quality database, which is one of the usual approaches today~\cite{LPIPS18}. 
Regardless of the task for which the deep-learning distortion metrics have been trained for~\cite{LPIPS18,Kumar22,Pablo_NeuralNets_25}, the common ground of the current stat-of-the-art metrics is their complexity compared to the simplicity of~\cite{Laparra17}, or other (relatively shallow) nonparametric measures which also include the bio-inspired end-to-end optimized DNs, as for instance, \emph{the PerceptNet}~\cite{Hepburn20}. 

As nowadays is very easy to find powerful regression tools, some people think that the \emph{image quality problem} is fully understood since it could be reduced to maximizing the correlation with a subjectively rated database~\cite{Nips11}.
However, this naive view may not be true~\cite{Martinez19}. 
We argue that the complexity (or absence of biological/statistical constraints) is an intrinsic problem of the conventional deep learning models. Networks with millions of parameters necessarily have robustness and generalization problems that are well known by the deep learning community. For instance, a deep learning method, LPIPS~\cite{LPIPS18}, does not outperform classical models in classical human-rated databases \cite{Pablo_NeuralNets_25}. A related problem is that detection thresholds of sensible visual features in conventional deep-perceptual metrics are far from human~\cite{Alabau24}.  

More importantly, conventional nonparametric deep metrics are hardly interpretable as vision models. For instance, it is hard to interpret the model features because these are spread along the layers of the nets in non-trivial ways. 
Preliminary evidences of that this \emph{feature-spreading} problem is critical were presented in conferences~\cite{Malo15,CIPJorge22,Malo22b}. 
In fact, Goodfellow et al. motivates the introduction of \emph{batch normalization} with a similar reasoning~\cite{Goodfellow16}: layers are individually updated despite they are functionally related. Using additional nonlinearities between linear layers (e.g. ReLUs + batch normalizations) certainly alleviates the mentioned parameter redundancy problem. However, it does not solve it completely.

In this paper, we propose a model that has the principles of vision science at its roots, but has been designed from the point of view of the deep learning paradigm. In particular, we propose a parametric neural network that can be viewed as a vision model. The functionality and default parameters of each of the layers are bioinspired. The network has 3~orders of magnitude fewer parameters than the nonparametric version, with less overfitting, more stable training, interpretable features, and comparable performance in regression tasks.
The model is based on a nonparametric approach, the \emph{PerceptNet}~\cite{Hepburn20}, because it was designed with the appropriate number of layers to accommodate the different known vision facts in the retina-V1 pathway. Moreover, it is equipped with bioinspired DN activations. From that, we parametrize \emph{all} the layers, dramatically reducing the model complexity, while keeping the architectures comparable.
This new setting allows an exploration that has not been done before: we study how, after initializing the parametric model with biologically plausible values, subsequent optimization can result in either biologically faithful or divergent outcomes depending on the degree of constraint imposed; this is what we call here the \emph{feature-spreading} problem. 
This opens new avenues for evaluating how closely such models align with human visual behavior and the need for novel evaluation metrics beyond traditional benchmarks. 

The structure of the work is as follows: 
Section~\ref{Structure} sketches the general structure of the \emph{Parametric PerceptNet}, and
shows a motivating example that illustrates the positive bias induced by the parametrization.
Section~\ref{sec:parametric_model} addresses the conceptual difference between parametric and nonparametric models in the training, and presents formally the model. After that, we describe the data (sec. \ref{sec:exp_setting}) used in the experimental Section (sec. \ref{sec:experiments}), where first, we analyze the effect of the bioinspired layers and the provided default parameters. We compare training of the parametric and the nonparametric models and analyze them in vision science terms for visualization and interpretability. Section~\ref{sec:discussion} discusses the reasons for the observed behaviors. And finally, section \ref{sec:conclusions} concludes the work and provides open questions. The Supplementary Material includes: 
A)~Functional expressions of the layers. 
B)~specific parametrization options. 
C)~ Psychophysically meaningful initialization. 
D)~Description of the illustrative image used in the experiments.
E)~Visualization of the stimuli that maximize the response in texture-sentitive layers.
F)~Possible functional improvements of the color and brightness front-end.

\section{Sketch of the Parametric PerceptNet}
\label{Structure}

\begin{figure*}[h!]
    \centering
    \includegraphics[width=1\textwidth]{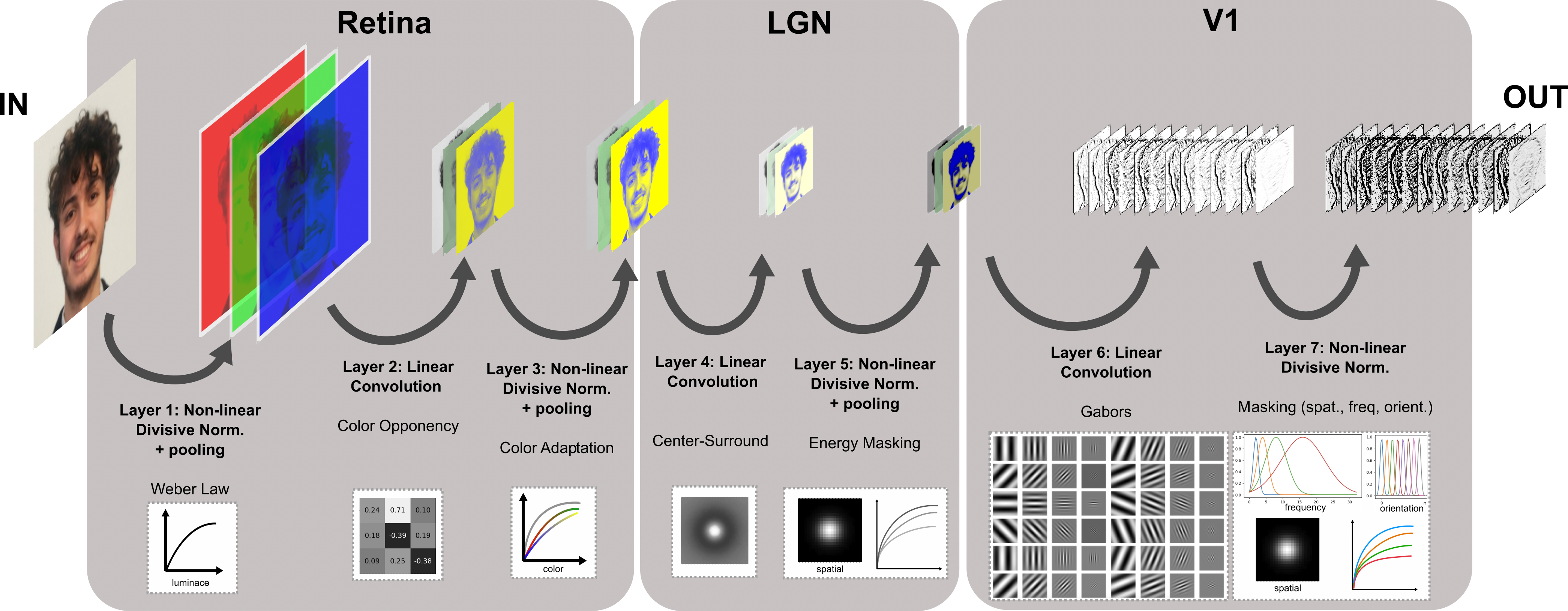}
\caption{\small{\textbf{Scheme of the proposed \emph{Parametric PerceptNet}}. The architecture is the same as in \cite{Hepburn20} but substituting each layer by its functional version. 
    The figure illustrates the widely known stages of the retina-LGN-V1 pathway: 
    inspired by visual neuroscience, it is a specific cascade of linear+DN layers as in~\cite{Martinez18,Malo24}.
    First, consistently with the Weber law, possibly happening in the LMS photoreceptors~\cite{Stiles00,Brainard10,Fairchild13} the first (nonlinear) stage includes a DN~\cite{Hillis05,Hillis07} in the input RGB channels. The second (linear) stage is meant to perform a transform to color opponent channels~\cite{Jameson57,Vila23}, the third (nonlinear) stage is meant to implement the adaptive nonlinearities of the RG-YB channels~\cite{Gegen92,Hita93,Laparra12,Laparra15}, the  fourth (linear) stage applies center-surround kernels to the (achromatic, red-green, yellow-blue) outputs of the previous stage following what is known of the LGN~\cite{DeAngelis97,Shapley11} and/or CSFs~\cite{Campbell68,Mullen85,Li92,Li22}. Layer five (nonlinear) is a DN that is meant to represent generic energy masking of whatever frequency (i.e. in the spatial domain)~\cite{Watson02,Malo15,Martinez18}. Layer six (linear) applies Gabor wavelets mimicking the shape of V1 receptive fields~\cite{Hubel59,Hubel61,Blakemore69,Watson90,Simoncelli92,Olshausen96}. Finally, the last layer implements a final DN to accommodate the known spatial-frequency-orientation masking that happens among Gabor channels~\cite{Ross91,Foley94,Heeger92,Watson97,Carandini12}. 
    At the V1 stages the linear gain of the Gabor filters (or scale of the final DN) is critical to keep the bandwidth of the system as described by the CSF~\cite{Malo97a}.
    All the activation-like nonlinearities are implemented using DNs as is natural in vision science \cite{Carandini12}. These \emph{non-standard} nonlinearities (unlike ReLUs or sigmoids) have parameters and are multidimensional~\cite{DN_in_enciclopedia19}. Therefore their responses are not a single curve but multiple curves.}}
    \label{fig:collage}
\end{figure*}

The general structure of the proposed model, the \emph{Parametric PerceptNet}, is illustrated in Fig.~\ref{fig:collage}. 
Although fairly standard from the vision science perspective (recent models such as~\cite{Laparra17,MantiukXX,MantiukYY,Martinez18,Martinez19,Gomez19,Malo24} have similar structure and physically/psychophysically calibrated flavor), the important point from the technical perspective is that \emph{its architecture is completely comparable} to the previously reported end-to-end optimized nonparametric \emph{PerceptNet}~\cite{Hepburn20}. The original \emph{PerceptNet} is a fair illustration of the current approach to build image quality metrics using deep networks. Note that it is very similar to Alexnet~\cite{Alexnet12} and VGG models~\cite{VGG16} in the sense of being a relatively shallow, feed-forward net with progressive reduction of resolution and increase of feature channels. However, \emph{PerceptNet} has a crucial \emph{biological} difference with those: the use of Divisive Normalization (DN) as activation function. In fact, DN is absent in VGG but kind of implemented in Alexnet. What Krizhevsky et al. call as local normalization of features (a form of DN)  is the second more important factor of its performance after ReLU~\cite{Alexnet12}.  In this regard, there are several reports~\cite{Ren17,Giraldo19,Giraldo21,Burg21,Martinez18,Hepburn20} that propose DNs as more biologically inspired than the one presented in~\cite{Alexnet12}. 

In contrast, the proposed \emph{Parametric PerceptNet} has a \emph{crucial difference} with regard to the original (nonparametric) \emph{PerceptNet}: instead of using  unconstrained kernels (both in the linear layers and in the DNs), here we enforce each layer to perform the mathematical operation inspired from a biological point of view. To do so, following Fig.~1, we define a parametric (functional) form for each layer (see Sect.~\ref{sec:model} and Supplem.~A and B). This will allow us to propose a set of bioinspired parameters (Supplem.~C). In order to enforce fair comparison, we will check also the effect of fitting the model parameters to maximize the alignment with subjectively rated image quality databases (as in the case of the original \emph{Perceptnet} or LPIPS). Training a parametric model has some specifics to take into account described in section \ref{sec:framework}. 

\subsection{Motivating example} % : \\ Parametrization-induced Bias
\label{sec:MotivatingIllustration}

\begin{figure}[t!]
    \centering
    \vspace{-0.5cm}
\includegraphics[width=0.5\textwidth]
    %,height = 0.2\textwidth]
    {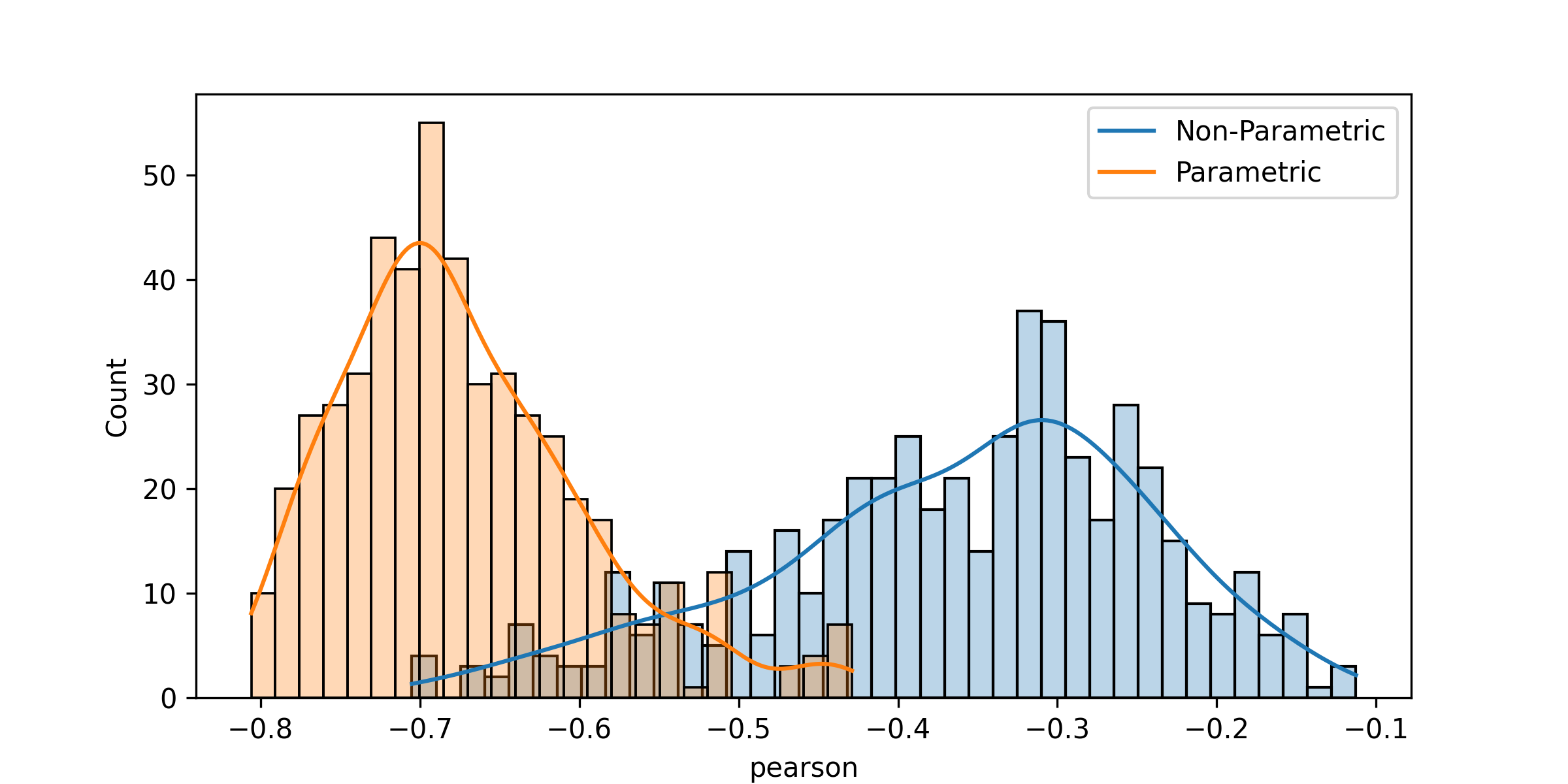}
\caption{\small{\textbf{Positive bias induced by the parametrization.} Histogram of Pearson correlations between 500 random initializations of the parametric (and the nonparamtric) \emph{PerceptNet} and human opinion (TID2008 \cite{ponomarenko_tid2008_2009}). Larger negative correlation means more aligned with human pinion.}}
\vspace{-0.5cm}
    \label{fig:compare_seeds}
\end{figure}

Before diving into the implementation details of the proposed \emph{Parametric PerceptNet}, we first illustrate the positive bias introduced by parametrization. Following standard practice in image/video quality evaluation, we use Pearson correlation with Mean Opinion Score (MOS) as the performance metric (see sec. \ref{sec:exp_setting}). We compare Pearson correlations results from multiple random initializations, without training, for both the nonparametric PerceptNet and the proposed parametric version. Fig.~\ref{fig:compare_seeds} shows performance histograms on TID2008. Throughout the work,  correlations are negative since higher distances correspond to lower MOS.

By looking at the distributions we can see that the Parametric PerceptNet is better since it attains bigger correlations, and
the distribution is sharper and peaks towards larger correlations. This result suggests advantages beyond the obvious reduction of parameters. First, the proposed parametrization indices a positive bias, which is not surprising given its inspiration in the referred vision science facts. Experiments below present explicit evidences that confirm this intuition. But also, as it is known~\cite{Kren22}, parametrization forms are easier to interpret since parameters have a particular physical meaning. This \emph{problem-aware} approach is not as commonly seen in the conventional deep-net approximation to image quality.

\section{Implementation Details}
\label{sec:parametric_model}

In this section we address three issues:
(A) the conceptual and technical difference between training models with unconstrained tensors and models in which tensors have to follow a certain functional form.  
(B) The specific 8-layer structure of the model, and 
(C) the issue of scaling of the linear and the nonlinear layers, which is important (as shown in the experiments below) both for the stability of the models during training and, more importantly, for their interpretablity.  

\subsection{Optimizing Parametric Models}
\label{sec:framework}

Deviating from the traditional approach in deep learning of optimizing directly the weights of the tensors (or convolutional kernels) used during the calculations, in this work we propose to optimize a series of generative parameters that are going to be used to calculate the whole weight tensors. 
By doing so, we can restrict our weight tensors to have a specific form of our liking, while reducing considerably the quantity of trainable parameters of our model. Even of greater interest is the fact that, by parametrizing the kernels of a convolutional neural network, we can independize the size of the convolutional kernels (both spatial and channel-wise) from the number of trainable parameters. This is impossible to accomplish within the traditional framework of optimizing each kernel weight.

\begin{figure}[t!]
    \centering
    \vspace{-0.5cm}
    \includegraphics[width=0.45\textwidth]{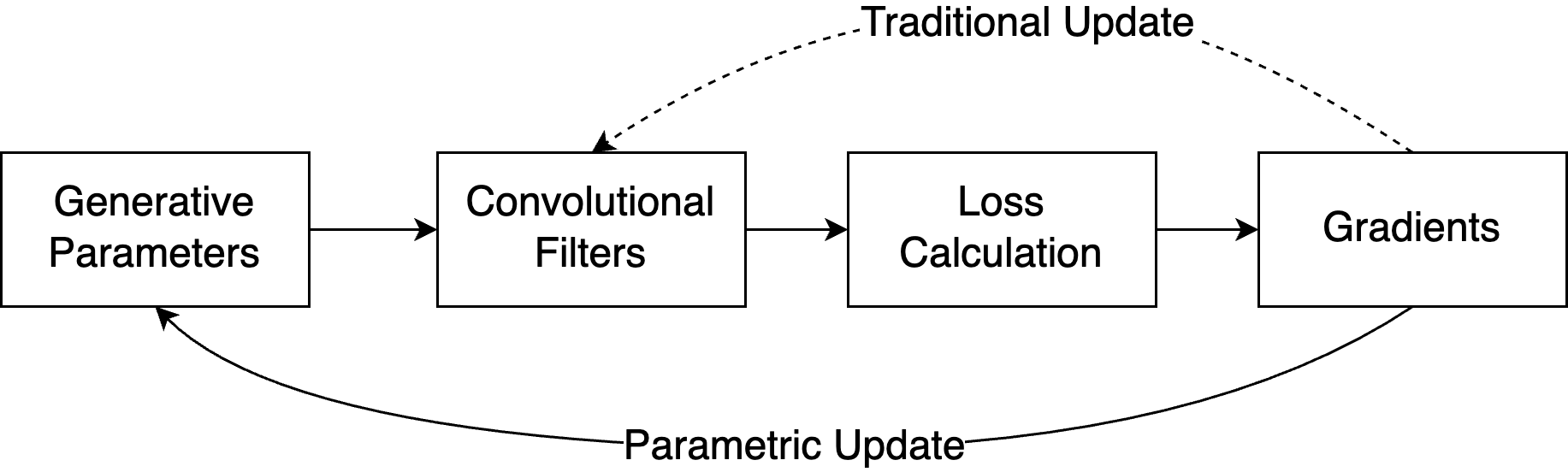}
    \caption{\small{\textbf{Illustration of an optimization step in a parametric layer.} The optimizable parameters are the generative parameters instead of the convolutional filters as a whole, which requires that we generate the convolutional filters at each step of the optimization so that the gradient can reach the generative parameters.}}
    \label{fig:framework}
\end{figure}

Not only does this approach reduce the number of trainable parameters in our model. The most notable feature is that it allows us to introduce prior knowledge into the network in the form of the parametrizations chosen. This can be specially interesting when defining models that reproduce already known behaviors, as is the case when modelling the visual system. It is known that the Retina-LGN-V1 path can be modeled by a series of transformations that include applying center-surround and Gabors filters (see expressions in Supplementary Material~A), both of which can be introduced into a network as parametric convolutions. Moreover, more biologically plaussible non-conventional activation functions can be employed. In our case we will use the divisive normalization \cite{Carandini12, Balle16}. 
By imposing specific functional forms, we avoid the classical trial-and-error approach in neural networks, where architectures are chosen and trained with the hope of reaching a solution with certain known form. Parametrizing convolutional kernels also enforces biologically inspired receptive fields and narrows the parameter search space.

This methodology introduces an extra step during training as the kernels have to be generated using the parameters, so that the gradient can reach them. While it can incur in an increase of computational time, this cost can be reduced by parallelizing the generation of the kernels. When performing inference with the model, these kernels do not have to be generated again, recovering the calculation speed of the non-parametrized models. A schematic representation of an optimization step within this framework is shown in Figure \ref{fig:framework}.

\subsection{The 8-layers structure of the Parametric PerceptNet}
\label{sec:model}

Here we describe the proposed model based on early vision, which is fully parametric, i.e. each layer has a parametric function. %Here we outline the function of each layer. 
We provide more explanation in the Supplementary Material: Suppl.~A (mathematical formulas), and Suppl.~B (implementation options).  %Whenever padding is needed, symmetric padding is used instead of 0s because it improves the performance. 
We will follow the architecture of the model shown in Fig. \ref{fig:collage}.

\begin{itemize}
\item {\bf Layer 1 (DN - "Weber adaptation")}: this layer mainly enhances low luminance ranges while compressing high luminance ranges following the idea in Weber law. In order to do so, we apply a divisive normalization independently to each channel and pixel (there is no interaction between pixels nor features).

\item {\bf Layer 2 (Convolution - "Opponent Color Space")}: it is meant to reproduce the transformation to color opponent space. A spatial max pooling of $2 \times 2$ is applied after the transformation.

\item {\bf Layer 3 (DN - "Krauskopf-Genenfurtner Adaptation")}: it produces color adaptation and is applied via a divisive normalization. It acts independently on the three channels, but the weights are not shared (as opposed to the first layer). There is no spatial interaction either.

\item {\bf Layer 4 (Convolution - "LGN center-surround cells")}: applies a set of Center Surround receptive filters as seen in retina and LGN~\cite{DeAngelis97}. This is implemented with a Difference of (Symmetric) Gaussians (DoG) (Supplementary Material A). We allow feature interaction in this layer, so we use 9 DoGs (3 input channels x 3 output channels) each with a different weight. The weights are chosen so that there is no interaction at initialization time. A spatial max pooling of $2 \times 2$ is applied.

\item {\bf Layer 5 (DN - "Generic energy contrast masking")}: replicates the LGN normalization by applying a divisive normalization, where its kernel is a parametric Gaussian, independently to each input channel.

\item {\bf Layer 6 (Convolution - "Gabor receptive fields in V1")}: visual cortex (V1) simple cells modeled by applying the convolution operation between a set of parametric Gabor filters and the input. Each output channel corresponds to a specific frequency and orientation, i.e. a different Gabor filter. We employ 
64~filters for the achromatic channel and 32~filters for each of the chromatic channels. The achromatic filters are combinations of 
4~frequencies, 8~orientations and 
2~phases, while the chromatic filters only have two possible frequencies. As in Layer~4, we allow feature interaction but the weights that regulate it are initialized so that there is no interaction.

\item {\bf Layer 7 (DN - "Space/frequency/Orientation masking in V1")}: divisive normalization in V1 that combines the outputs of the simple cells from the previous layer. As the previous layer's filters are Gabors, we can define the denominator kernel so that there is a Gaussian relation between input features, taking into account not only the spatial position but also the frequency and orientation of the features. The exact expression can be found in Supplementary Material~A.

\item {\bf Layer 8 (Final Linear Scaling)}: This is a special linear layer included to allow the model to adapt to each particular task, in our case adapted for image quality.  

\end{itemize}

As a result, the described model has a total of 1062 parameters. In Suppl.~C we provide a set of bioinspired parameters for this layers that are in agreement with the psychophysics and physiology evidences described in Sect.~\ref{Structure}.

\subsection{On the scaling of linear and nonlinear layers}
\label{scaling}

Control of the \emph{scaling} of the linear and nonlinear stages is key in order to understand what we call \emph{feature spreading problem} mentioned in the introduction. This will become apparent later in the discussion of the results. In principle the scaling (or normalization) of \emph{all} layers (both linear and nonlinear) should be explicitly controlled and be separately frozen or optimized if necessary.
This is intuitively obvious in the linear cases: note that concatenation of linear layers naturally introduces a competing uncertainty between the layers. The training may get trapped in a meaningless oscillation where one of the layers is amplified and the other is attenuated while not changing anything crucial (e.g. the shape of their frequency responses). The scale of the nonlinear layers has to be controlled in the same way for the same reasons.

In the \emph{Parametric Perceptnet} we made particular choices to control this scaling, particularly in the layers that, in principle, seem more important from the texture point of view: the scale of the center-surround neurons, the Gabor neurons and the weights of the nonlinear DN after the Gabors.
Imposing normalization constraints for the linear (convolutional) layers is trivial and a general way for the DN transform has also been proposed~\cite{Martinez19}. However, we did not systematically apply those constraints for consistency with the original nonparametric PerceptNet, where the scale of the DNs is not controlled in any way. Regardless of the specific scaling choices made, detailed in Supplementary Material~A and B, the experiments will show the critical implication of this kind of choices and how important they are to have explainable visual features in the models for image quality assessment. 

\section{Experimental setting}
\label{sec:exp_setting}

The numerical results in this paper are given following the classical approach for evaluating image quality metrics. We measure the correlation between the distortions predicted by the model and the subjective distortions reported by human observers when rating the quality of a database of distorted images~\cite{Wang08,Wangbook06,Watson02,Martinez19}. The deep learning community used this classical evaluation approach as a loss function to train the models~\cite{Laparra17,LPIPS18,Hepburn20}. 

We are going to consider 3 different databases used as follows: TID2008 \cite{ponomarenko_tid2008_2009} will be used for training our model, TID2013 \cite{ponomarenko_image_2015} will act as a validation set, and KADID \cite{kadid10k} will act as test set. We chose this configuration because TID2008 and TID2013 share the same reference images but have different distortions, so validating with TID2013 allows us to assess the distortion generalization capabilities of our model. On the other hand, KADID is chosen as test set because it has different reference and distorted images, so it can be considered as a good approximation to the image and distortion generalization capabilities of the model when deployed into the real world. Keep in mind that we will be using the smallest dataset to train our model, which is also interesting on its own because being able to obtain good generalization performances with as little training data as possible is always a thing to look out for.

Since these databases are collected among many observers and under loosely controlled observing conditions, each subjective rating has a lot of variance.
In short, conventional databases are very useful but overfitting to them has no real point~\cite{Martinez19}. Therefore, it is important to have in mind the self consistency of the databases. Table~\ref{tab:iqa_datasets} presents an estimation of the self consistency of these databases estimated with a simple Monte Carlo experiment. We sampled from a normal distribution with the mean and the variance of each subjective rating provided by the database\footnote{The normal distribution assumption is just an option to do sampling. The actual distribution (which is not provided in the databases) may be non-Gaussian, and this would change the values a bit for the maximum attainable correlation, but the general conclusions would be qualitatively the same.}. This simulates responses from different observers for the whole database, and we compute correlations (in particular Pearson $\rho$) between these "simulated observers". It is not reasonable that any metric should get a correlation bigger than the upper bound of these correlations ($\rho_{\textrm{max}}$). Note that state-of-the-art metrics tuned to maximize the correlation usually obtain larger correlations than these estimated $\rho_{\textrm{max}}$. 

\begin{table}[h]
\caption{\small{Size and self-consistency of some popular databases.}}
\footnotesize
\label{tab:iqa_datasets}
\centering
\begin{tabular}{|c|c|c|c|c|c|}
\hline
Dataset & \# Samples & \# Refs & \# Dists & \# Intensities  & $\rho_{\textrm{max}}$ \\
\hline
TID2008 & 1700 & 25 & 17 & 4 & 0.86\\
TID2013 & 3000 & 25 & 24 & 5 & 0.83\\
KADID & 10125 & 81 & 25 & 5 & 0.78\\
\hline
\end{tabular}
\end{table}
\section{Experiments and Results}
\label{sec:experiments}

In this section, we first analyze the effects of the bioinspired layers and parameters. On the one hand, we perform an ablation analysis layer by layer of the differences when using non-parametric layers or parametric bioplausible layers (sec. \ref{sec:relevance}). On the other hand, we analyze the effect of fitting the parameters to maximize correlation or directly setting them to be bioplausible (sec. \ref{sec:dist_to_optim}). 

After that, we analyze three versions of the model described in \ref{sec:models}: one fully non-parametric, and two fully parametric. One of the parametric with the parameters restricted to be bioplausible and the other one fitting the parameters to maximize correlation. We evaluate the three models in different terms. We compare their performance with classical models, and analyze their behavior during training (sec. \ref{sec:training}) and, we interpret the three models in vision science terms by analyzing the performed transformations and the model features (sec. \ref{sec:interpretability}). 

\subsection{Relevance of bioinspired layers}
\label{sec:relevance}

In order to assess the effect of substituting the classical non-parametric layers by the parametric bioinspired, we perform an ablation study where we compare identical models with and without parametrization at different layers. We tested all the possible combinations of layers that differ from the non-parametric model. Note that we can only parameterize the V1 normalization (layer 7) if we also parameterized V1 Gabor (layer 6) because the normalization needs to know the frequencies and orientations of its inputs and this information is only provided by the parametrized V1 Gabor layer. We use the same kernel size for both the parametric and non-parametric models to keep the comparison as fair as possible. %This results in a non-parametric model with more trainable parameters and thus more learning capabilities. 

Results in table \ref{tab:res_pearson} show that, in general, introducing parametric variations at different stages reduces the number of parameters, but does not hinder its performance. With the evaluated parameterizations we are able to obtain models with only 5k parameters that perform better than a model with more than 7.5M. The most notable reduction in parameters comes from parametrizing the V1 Gabor and V1 normalization layers. All models are fitted to maximize the correlation in TID2008, and perform above the human-like correlation (Table \ref{tab:iqa_datasets}) for both datasets. Interestingly, all the models that perform better than the non-parametric version include the parametrized V1 Gabor layer. This shows that enforcing Gabor receptive fields at this stage of the network is a reasonable bias and improves performance. 

As opposed to this, including the parametrized DoG layer (layer 4) results in a decrease in performance. This may point to an interaction between the DoG and V1 Gabor layers (a special case of the \emph{feature-spreading problem}). %that should be given special attention in further work. 
Center-surround receptive fields represent circularly symmetric filters that may be band-pass or low-pass in the Fourier domain, but combinations of Gabor filters include those possibilities too. This redundancy that is certainly present in the visual system~\cite{DeAngelis97,Shapley11,Hubel59,Hubel61} may have a negative effect. The existence of center-surround cells in the LGN may be due to \emph{other reasons}: a number of authors have argued that these cells are there to enhance the retinal signal~\cite{Li92,Molano14,ICLR19,Li22}. 

Contrary to intuition, the model with more parameters is not the model that performs better in either training or validation. This is an evidence that having too many parameters in the non-parametric model negatively affects the training process. Optimizing the last linear scaling (layer 8) results in a performance increase, since it helps to adapt the outputs to the particular task.

\begin{table}[h]
\caption{\small{Ablation analysis changing the non-parametric layers by their parametric version, pearson correlation is shown. Numbers in brackets indicate which layers are parametrized. Models that perform better than the Non-Parametric version are in bold.}}
\label{tab:res_pearson}
\centering
\begin{tabular}{|cccc|}
\hline
Param. Layers & TID2008 & TID2013 & \# Params \\
\hline
$[1]$ & -0.95 & -0.89 & 7598848 \\
\hline
$[1,4]$ & -0.94 & -0.90 & 7594910 \\
$[1,5]$ & -0.94 & -0.90 & 7598495 \\
$[1,6]$ & \textbf{-0.97} & \textbf{-0.91} & 7230280 \\
$[1,8]$ & -0.96 & -0.90 & 7598976 \\
\hline
$[1,4,5]$ & -0.95 & -0.90 & 7594549 \\
$[1,4,6]$ & -0.94 & -0.89 & 7226342 \\
$[1,4,8]$ & -0.95 & \textbf{-0.91} & 7595038 \\
$[1,5,6]$ & -0.94 & -0.90 & 7229927 \\
$[1,5,8]$ & -0.94 & -0.90 & 7598623 \\
$[1,6,7]$ & \textbf{-0.96} & \textbf{-0.91} & 5233 \\
$[1,6,8]$ & -0.95 & -0.90 & 7230412 \\
\hline
$[1,4,5,8]$ & -0.94 & -0.90 & 7594677 \\
$[1,4,6,8]$ & -0.95 & -0.89 & 7226470 \\
$[1,4,5,6]$ & -0.93 & -0.89 & 7225985 \\
$[1,4,6,7]$ & -0.93 & -0.89 & 1295 \\
$[1,5,6,7]$ & -0.96 & -0.90 & 4880 \\
$[1,5,6,8]$ & \textbf{-0.97} & -0.90 & 7230055 \\
$[1,6,7,8]$ & -0.96 & \textbf{-0.91} & 5365 \\
\hline
$[1,4,5,6,7]$ & -0.93 & -0.89 & 938 \\
$[1,4,5,6,8]$ & -0.94 & -0.89 & 7226113 \\
$[1,4,6,7,8]$ & -0.93 & -0.90 & 1423 \\
$[1,5,6,7,8]$ & \textbf{-0.98} & -0.90 & 5008 \\
\hline
$[1,4,5,6,7,8]$ & -0.93 & -0.89 & 1062 \\
\hline
Non Parametric & -0.96 & -0.90 & 7598852 \\
\hline
\end{tabular}
\end{table}

\subsection{Effect of the bioinspired parameters}
\label{sec:dist_to_optim}

\setlength{\tabcolsep}{4pt}
\begin{table}[t]
\centering
\caption{\small{Performance of the model when freezing the initialization with bio-inspired parameters up to different depths in the architecture. Numbers in brackets indicate the frozen layers.}}
\label{tab:freezing}
\begin{tabular}{|c|c|c|c|c|c|}
\hline
Frozen layers & TID2008 & TID2013 & KADID & \# Param. & \makecell{\# Trainable \\ Parameters} \\
\hline
Fully Trained & -0.93 & -0.88 & -0.80 & 1062 & 1062 \\
$[1]$ &   -0.92 & -0.88 & -0.77 & 1062 & 1060 \\
$[1-2]$ & -0.91 & -0.88 & -0.79 & 1062 & 1051 \\
$[1-3]$ & -0.89 & -0.86 & -0.78 & 1062 & 1045 \\
$[1-4]$ & -0.89 & -0.86 & -0.75 & 1062 & 1018 \\
$[1-5]$ & -0.86 & -0.84 & -0.73 & 1062 & 1009 \\
$[1-6]$ & -0.85 & -0.82 & -0.71 & 1062 & 553 \\
$[1-7]$ & -0.83 & -0.81 & -0.71 & 1062 & 128 \\
$[1-8]$ & -0.42 & -0.46 &  -0.53 & 1062 & 0 \\
\hline
\end{tabular}
\end{table}
\setlength{\tabcolsep}{6pt}

In this experiment, we analyzed the difference between using hand-crafted bioinspired parameters or training them to maximize correlation. This experiment is performed using the model in which all the layers are bioinspired and parametric. 
%It is important to remind the reader that producing an interpretable handcrafted initialization is only possible because the model has a reduced set of easily interpretable parameters.
We initialize the model with the bioinspired parameters and train different versions of it where we only let the optimization algorithm change the parameters from a particular layer onwards. %The initializations are chosen assuming that the previo us layers behave in a particular way, so we won't be trying different combinations as in the previous experiment.
Results in Table \ref{tab:freezing} show what it is to be expected: the more freedom we give the algorithm to change the parameters, the better performance in correlation we get. However, this comes at the cost of obtaining not so understandable parameters (sec. \ref{sec:interpretability}). The most important take-away is that, even in the most restricted scenario (i.e. only fitting 128 parameters), the model performance is very close to the human performance bound (Table \ref{tab:iqa_datasets}). This indicates that the proposed bioinspired parameters are a good choice. %This supports the idea that human-like performance can be attained when constraining the parameters to be human-like but may differ considerably when pushing the model's performance into super-human territory. 

\subsection{Models selected for further analysis}
\label{sec:models}

\begin{table*}[h!]
\centering
\caption{\small{Pearson correlation of the proposed models, and state-of-the-art perceptual quality metrics. ($^{*}$) model was trained in this dataset.}}
\label{tab:res_final_models}
\begin{tabular}{|c|c|c|c|c|c|}
\hline
Model & TID2008 (Train) & TID2013 (Validation) & KADID (Test) & \# parameters & \# trainable parameters \\
\hline
Non-Parametric & \textbf{-0.96$^{*}$} & \textbf{-0.90} & -0.76 & 7.598.852 & 7.598.852 \\
Param. Fully Trained & -0.93$^{*}$ & \textbf{-0.89} & -0.80  & 1062 & 1062 \\
Param. Partially Trained & -0.83$^{*}$ & -0.81 & -0.71 & 1062 & 128 \\
\hline
SSIM & -0.65 & -0.72 & -0.66 & 3 & 3 \\
LPIPS & -0.70 & -0.73 & -0.70 & 14.714.688 & 1472 \\
DISTS & -0.80 & -0.83 & \textbf{-0.86$^{*}$} & 14.714.688 & 2950\\
\hline
Max. Exp. Corr. & -0.86 & -0.83 & -0.78 &  & \\
\hline
\end{tabular}
\end{table*}

Given the good results in correlation obtained by all the models, we select three configurations to be further analyzed in the following.
The selection represents the path between two extremes: the current conventional unconstrained approach, and the classical frozen extreme where all parameters are taken from a range of disconnected experiments of the visual neuroscience literature. Namely,

\begin{itemize}
    \item \textbf{\emph{Non-Param}:} Non-Parametric Perceptnet.
    All the layers are non-parametric, similar to the original PerceptNet \cite{Hepburn20}. The only thing that changes is the size of the convolutional filters to match the size of the parametric ones. The model is fully trained to maximize correlation with MOS in TID2008. 
    \item \textbf{\emph{Param-Fully}:} Parametric bio-initialized and fully trained.
    Model with all the layers parametric and with the parameters of all layers initialized to be biologically plausible (using values in Supplementary Material~C) but, \emph{all} parameters are fitted to maximize correlation with MOS in TID2008. 
    \item \textbf{\emph{Param-Partial}:} Parametric bio-initialized but partially optimized.
    Model with all the layers parametric and with the parameters of all layers initialized to be biologically plausible (using values in Supplementary Material~C) as above, but then only the final linear scaling (layer 8) is optimized to maximize correlation with MOS. In this way, we ensure that the model reasonably behaves according to the classical literature on human vision.  
\end{itemize}

Table \ref{tab:res_final_models} shows the performance of the models in different image quality databases. Note that all the three models have a similar or better performance in validation or test than the classical and the deep-learning based models.

\subsection{Behavior in training: stability and overfitting}
\label{sec:training}

\begin{figure}[b!]
    \centering
    \includegraphics[width=0.46\textwidth]{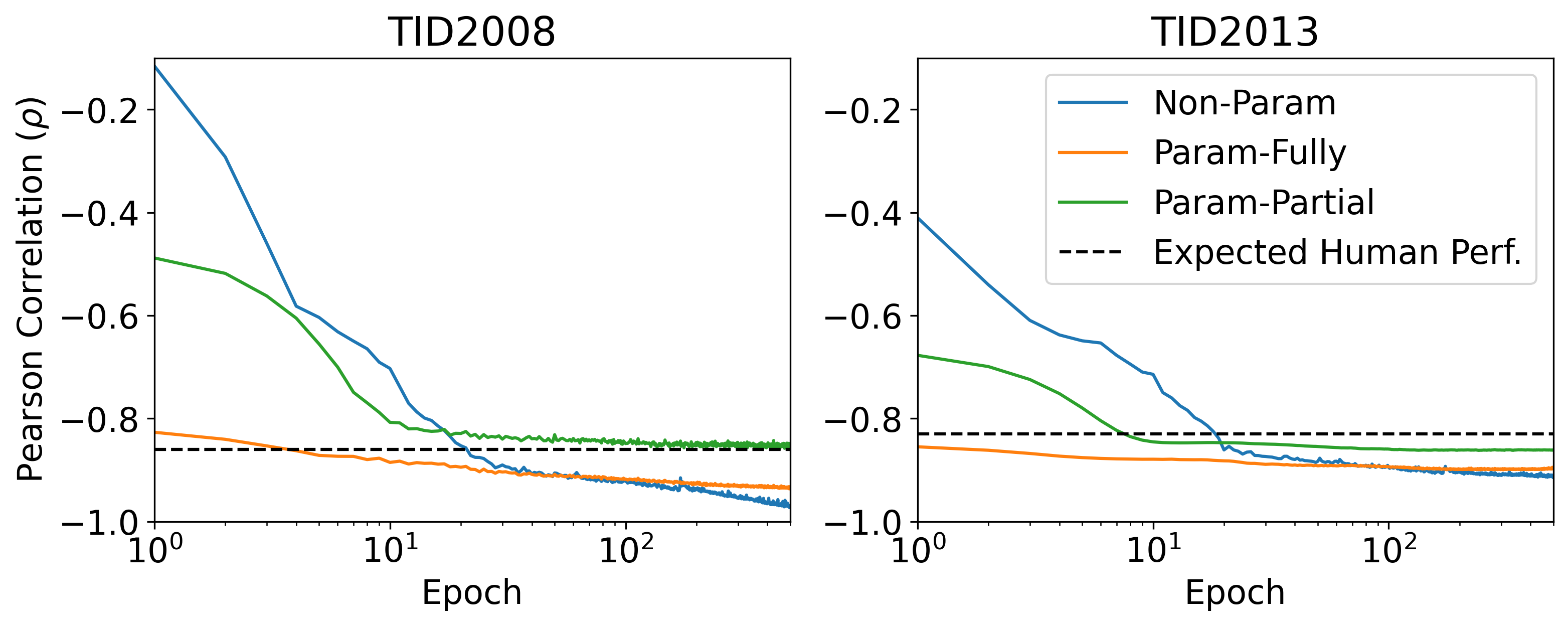}
    \caption{\small{\textbf{Learning curves for the three selected models.} Training is performed in TID2008 while TID2013 is used for validation. Results in test can be seen in Table \ref{tab:res_final_models}}. Note that both parametric models are initialized in the same way. The first point of each curve is calculated after the first epoch of optimization.}
    \label{fig:learning_curves}
\end{figure}

Figure \ref{fig:learning_curves} shows learning curves for the three models (sect.\ref{sec:models}). Parametric models outperform the non-parametric one by: 1) starting with higher initial correlation, 2) training faster (arrive earlier to the human performance), 3) showing smoother curves, and 4) better regularization (similar train and validation behavior).
This demonstrates that the proposed bioinspired parameters help in the initialization, and the chosen bioinspired parametric layers act as a regularization that potentially smooths the optimization landscape. In fact, both parametric models obtain human performance with just one epoch. Note that the Param-Partial model only has to train the last linear layer to adjust the importance of each channel (128 params). This could be done analytically using least-squares, but here we train it using gradient descent for fair comparison. 

The non-parametric model takes much longer to converge and overfits, which is easily visible in Table \ref{tab:res_final_models}, where it obtains much bigger correlation than the maximum experimental one in train and validation, but the performance is reduced in test. The parametric fully trained model gets similar results as the non-parametric while having 3 orders of magnitude less parameters. However, in Sec. \ref{sec:interpretability} we will see that it converges to a set of parameters that are not as human-like as one could expect. On the other hand, the param-partial model obtains a slightly lower performance but suffers less from overfitting. The training process is smoother and faster than the other models, and the results are fully interpretable as we will see in Sec. \ref{sec:training}.

\subsection{Experiments for feature visualization and interpretability}
\label{sec:interpretability}

In this subsection we analyze the biological plausibility of the selected models (Sec. \ref{sec:models}) with two experiments: 
First, we visualize the responses to an illustrative image and discuss whether the models capture known facts of low-level psychophysics that occur in the retina-V1 pathway. Then, we go through all the layers and visualize their receptive fields, features and nonlinear responses. 

\subsubsection{\textbf{Responses to an illustrative natural-synthetic image}} Here we analyze the response of the models to an image designed to illustrate different effects: opponent colors, textures with different orientations and frequencies, and a synthetic shadow to induce a spatial variation of contrast. A detailed description of the image and its generation is available in the Supplementary Material D.

% FIG A
\begin{figure*}[h!]
    \centering
\vspace{-0.5cm}
    \includegraphics[width=1\textwidth]{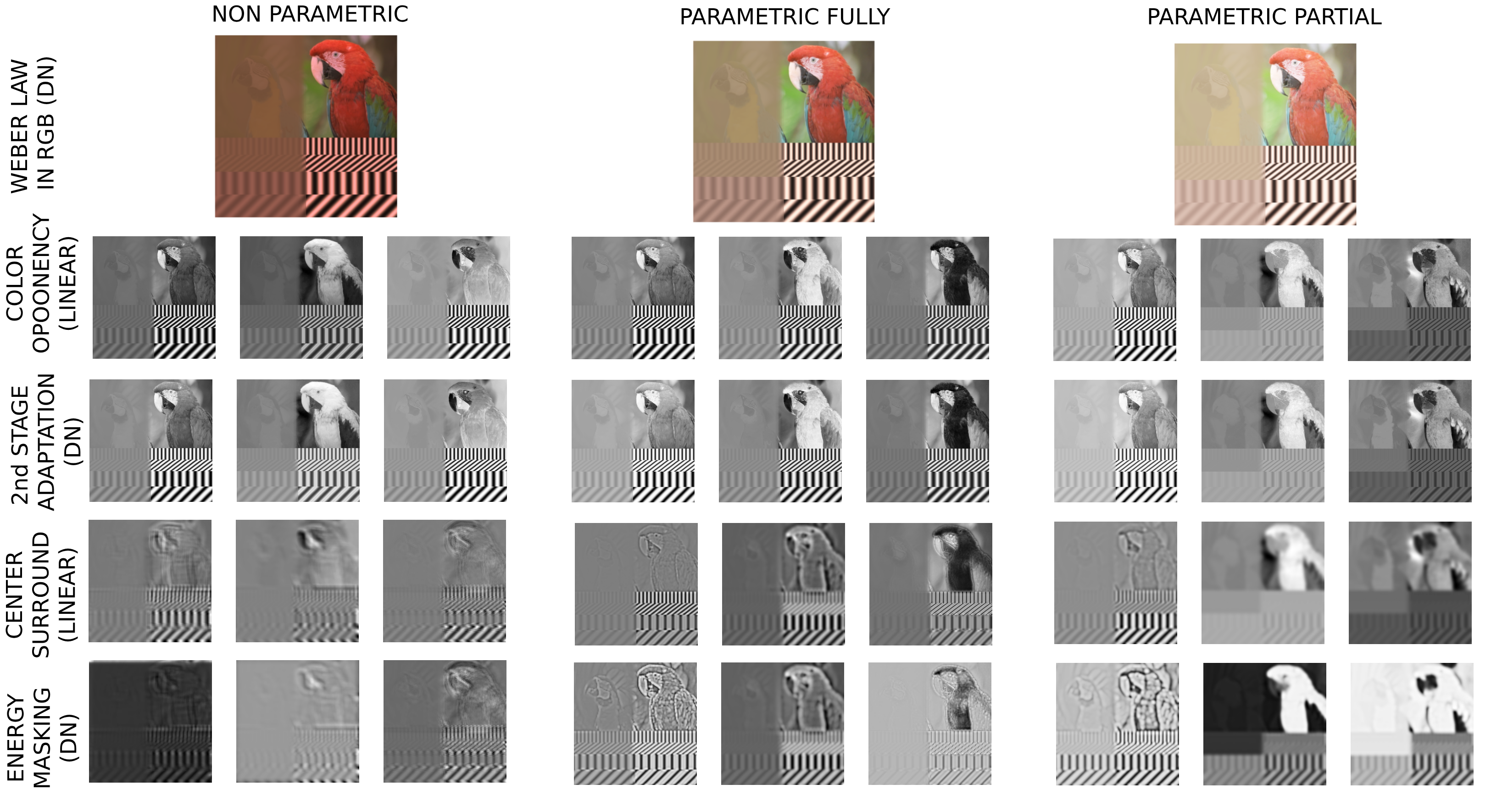}
    \caption{\small{\textbf{Illustrative image in the first layers (Retina-LGN),} in processing order (top to bottom), for the three models (left to right).}}  %Each macro-column represents the results for each model: \emph{Non-Param}, \emph{Param-Fully}, and \emph{Param-Partial}.}}
\vspace{-0.5cm}
    \label{fig:A}
\end{figure*}

Figure~\ref{fig:A} shows the response to the image of each layer before V1 for each model. The model that best reproduces the luminance-brightness saturating non-linearity (Weber law~\cite{Stiles00,Brainard10,Fairchild13}) is the Param-Partial model. Brightness in darker regions increases according to the known adaptation in humans, and color appearance (hues and saturations) is not modified. However, in the Param-Fully model, although there is a certain brightness equalization, it is less pronounced. The non-parametric model shows no biologically significant characteristics: the relationship between the scales among the color channels has become unbalanced leading to a color appearance change, and the image luminance has not been equalized. 

Moving into the second row, the Param-Partial model shows an opponent transformation~\cite{Jameson57,Vila23}: the leftmost image corresponds to the luminance, the central one corresponds to the red-green channel, as can be seen in the parrot, and the right one corresponds to the blue-yellow channel. However, in the non-parametric and Param-Fully models, the receptive fields of the opponent transformation have been deformed in such a way that their meaning is not biologically plausible. For all three models the following adaptation stage (third row) behaves similarly. It is initialized as a linear transformation, and is almost not changed during training. 

The next stage applies center-surround neurons. In the Param-Partial model, chromatic channels show low-pass behavior (strong blurring), while the luminance image shows band-pass behavior with edge enhancement. Low-frequency patterns are preserved, but high-frequency contrast is reduced. In red-green and blue-yellow images, high-frequency patterns vanish due to low cut-off frequencies, matching human CSFs~\cite{Campbell68,Mullen85,Li92,Li22}. %At the same time, we can observe that the cut-off frequency of the achromatic contrast sensitivity function is much higher than the chromatic ones.
Although Param-Fully and Param-Partial share the same initialization, Param-Fully shows diverged frequency behavior, with all periodic patterns similarly attenuated. Finally, the non-parametric model also shows an arbitrary bandwidth distribution across the three channels, lacking clear chromatic meaning.

Finally, the non-linear stage applies contrast enhancement in low contrast spatial regions. This is observed in the periodic patterns that have low contrast in the left part of the image for the Param-Partial model. This biological behavior~\cite{Watson02,Malo15,Martinez18} is not observed with the same intensity in the Param-Fully model, and not at all in the non-parametric one.

% FIG B
\begin{figure*}
    \begin{centering}
    \vspace{-0.5cm}\includegraphics[width=1\textwidth]{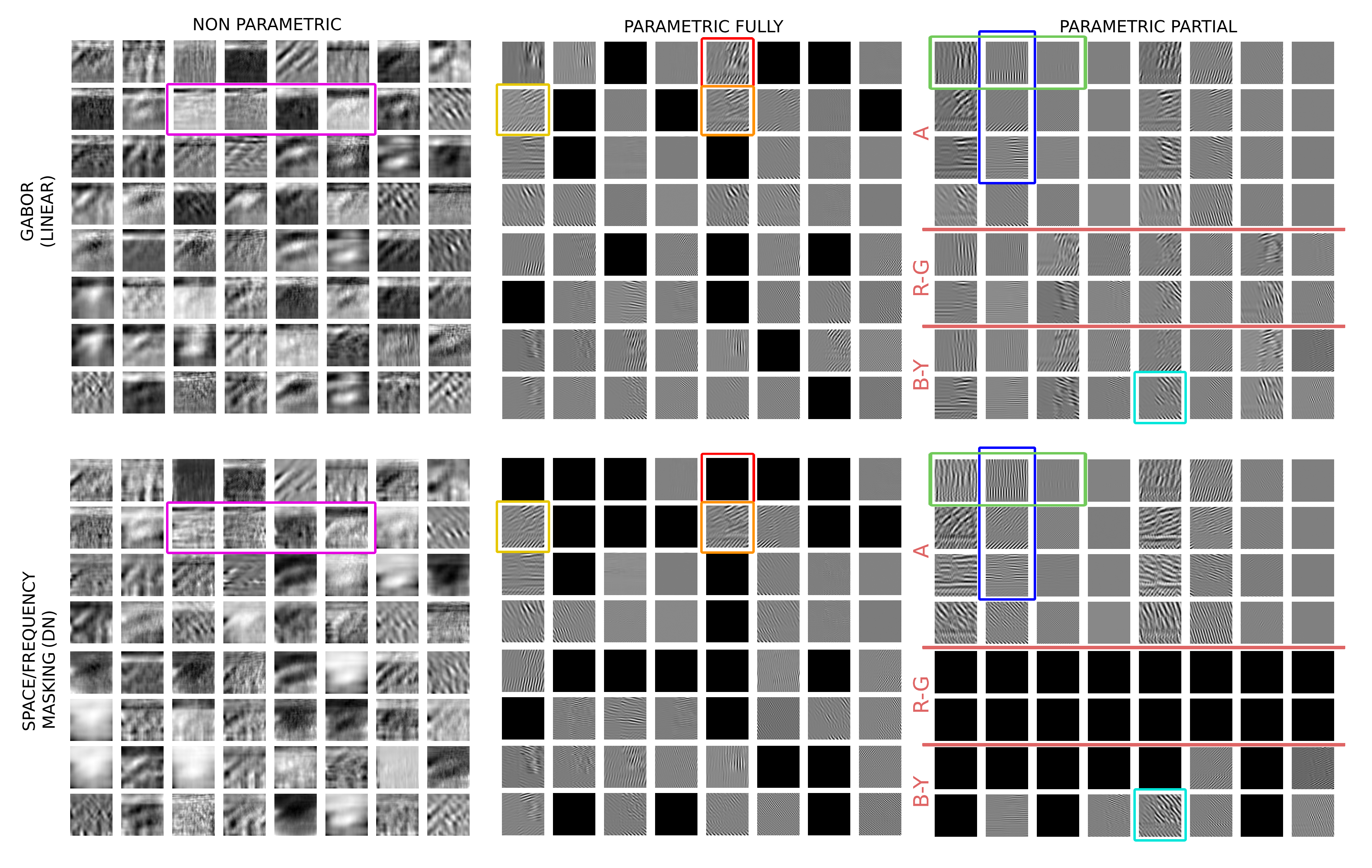}
    \end{centering}
    \vspace{-0.5cm}
\caption{\small{\textbf{Illustrative image represented in the last layers (linear-V1 and nonlinear-V1)}: after the 3rd convolution in Layer~6 (top) and after the final DN (V1 Normalization) with its final scaling (bottom). Only the first 64 of 128 outputs are represented for proper visualization. Each of the three macro-columns represents the results for each of the three models: non-parametric, parametric fully fitted, and parametric partially fitted. In the case of the non-parametric model the outputs does not resemble to the outputs of orientation and frequency selective filters, while in the parametric ones this is imposed (and achieved). In the case of the parametric trained this is more diffused and actually some outputs are discarded.}}
\vspace{-0.0cm}
\label{fig:B}
\end{figure*}

The behavior for V1 (last two layers) is shown in Figure~\ref{fig:B}. %The first 16 sets of the responses of the frequency and orientation sensitive neurons of the models are shown, both before divisive normalization in the upper panels and after divisive normalization in the lower panels.
In the case of the Param-Partial model, the responses highlighted in blue and green are interesting because they are tuned to different frequencies and orientations. For example, the square in green shows a strong response to vertical-type patterns of different frequencies, and the square in blue shows that each of these filters responds to patterns of the same frequency but different orientation.

The effect of divisive normalization consists of enhancing the response in regions where the input has small amplitudes corresponding to small contrasts \cite{Carandini12, Martinez19}, something that is seen in how the oscillations of the responses increase. This behavior, which is biologically plausible~\cite{Ross91,Foley94,Watson97} and observed in the Param-Partial model, is not observed at all in the Param-Fully model. For example, although the regions marked in red and orange do respond to specific frequency patterns, it is observed that after divisive normalization the scale of the responses does not make much sense: the band highlighted in red has completely disappeared while, in the band highlighted in orange, this contrast enhancement is not as intense as in the Param-Partial case, as can also be seen in the sub-band highlighted in yellow. 
The final scale in the Param-Partial model basically removes the channels with RG meaning: we will come back to the effect of this scale (the only tuned parameter in the Param-Partial model) later in Figs.~\ref{fig:gabor_resp}-\ref{fig:F} below.
Finally, in the non-parametric model, the linear filters have no particular physical meaning, therefore, the responses have no qualitative sense, and the divisive normalization does enhances oscillations but with no relation to the variable contrast in the input, as seen in the illustrative examples highlighted in pink.
The Supplementary Material~E shows the stimuli that maximize the response of the texture-sensitive layers represented in Fig.~\ref{fig:B}.

\subsubsection{\textbf{Visualization of receptive fields and nonlinearities}}

% FIG C
\begin{figure*}
\vspace{-0.0cm}
    \centering
    \includegraphics[width=1\textwidth]{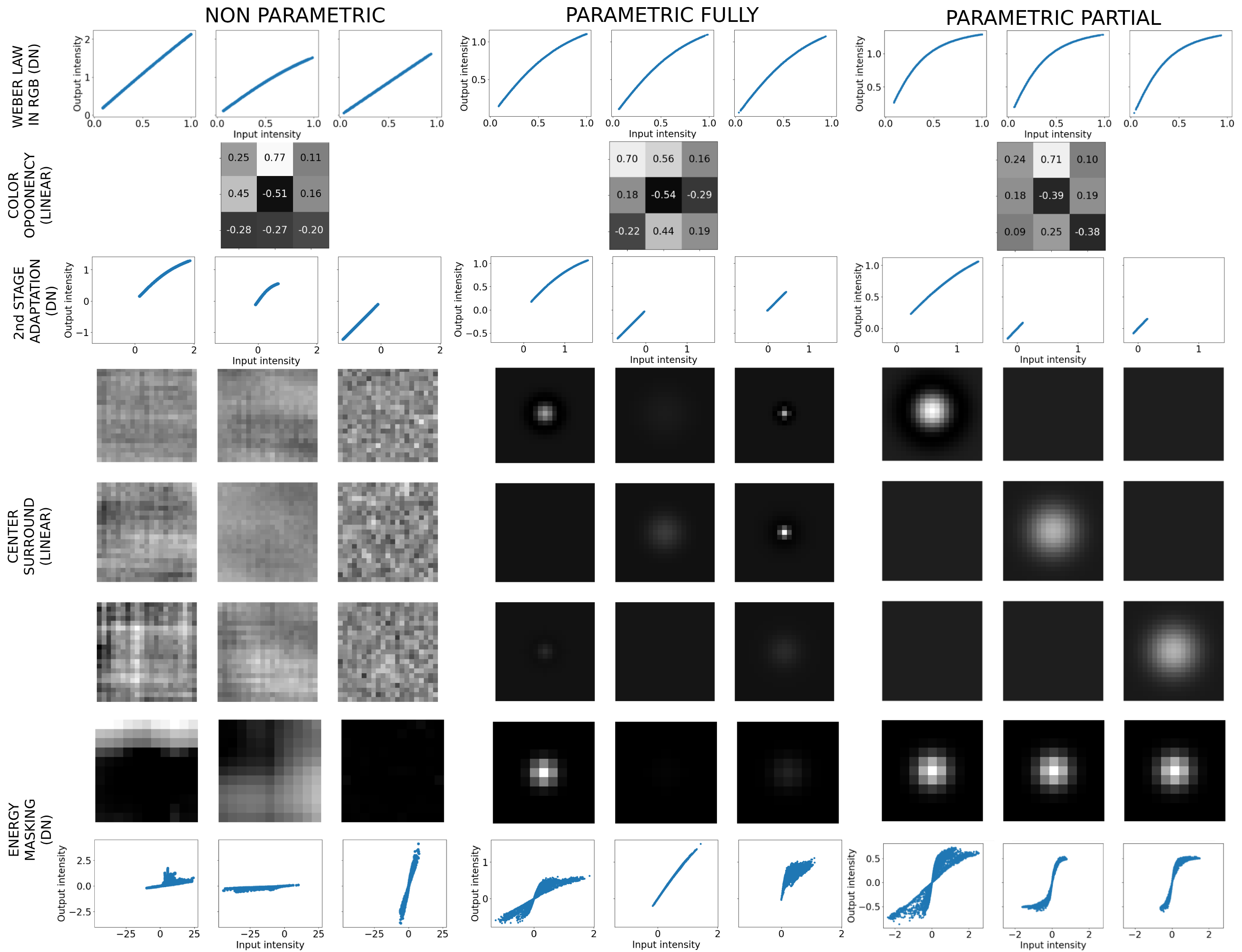}
\vspace{-0.65cm}    \caption{\small{\textbf{Input-Output scatter plots and Parameters (or kernels) for the Retina-LGN layers}. In each row we can see the effect of the layer over data (natural-synthetic image). In the non-linear stages a input-output function over a natural image (the one from figure \ref{fig:A}) is shown. For the linear stage the convolutional filters are shown. The transformations performed by the bio-fitted model are consistent with the literature. However, it is not the case for the other two.}}
    \vspace{-0.5cm}
    \label{fig:C}
\end{figure*}

The input and output responses observed in Fig.~\ref{fig:C}, as well as the receptive fields, are consistent with the type of responses represented in Fig. \ref{fig:A}. Specifically in the first layer, the Param-Partial model presents the more saturated input-output curves consistently with the Weber law~\cite{Stiles00,Brainard10,Fairchild13}. However, we see smoother saturation in the Param-Fully model, indicating that this non-linearity may not be relevant to solve the training set. Finally, in the non-parametric model, the non-linearity has almost completely disappeared and the scaling between the different RGB channels does not respect the relative scaling, producing a modification of the chromatic information.

In the matrix to change from RGB space to what should be opponent channels~\cite{Jameson57,Vila23}, it is seen that only in the Param-Partial model the biologically plausible transformation is enforced. In contrast, these matrices in the Param-Fully model are biologically incorrect: giving more relative relevance to the red channel, the central channel being basically an $R - G - B$ channel, when it should be $R - G$, and the last one being $G + B - R$, which also does not correspond to the blue-yellow channel. In the non-parametric model, although the first achromatic channel is correct, the other channels are arbitrary. The second stage of adaptation in the Param-Fully and Non-param models does not have a clear interpretation either. Regarding the center-surround receptive fields~\cite{DeAngelis97,Shapley11}, the non-parametric obtains filters far from being center-surround. In the Param-Fully case, despite the bio-inspired initialization, it diverges to a solution that is difficult to interpret.

With respect to the energy masking step, the Param-Partial model has a saturating and adaptive behavior with respect to the positive and negative values of the input, which has a very adequate biological interpretation~\cite{Watson02,Malo15,Martinez18}. In the Param-Fully model, the first and third channels show saturation, in agreement with biology. However, the third channel shows only positive responses and central channel has a more linear behavior, which do not make biological sense. The non-parametric case, effectively, makes no sense either.

% FIG GABORS
\begin{figure*}[t]
\vspace{0cm}
\centering
\hspace{0cm}\includegraphics[width=1\textwidth]{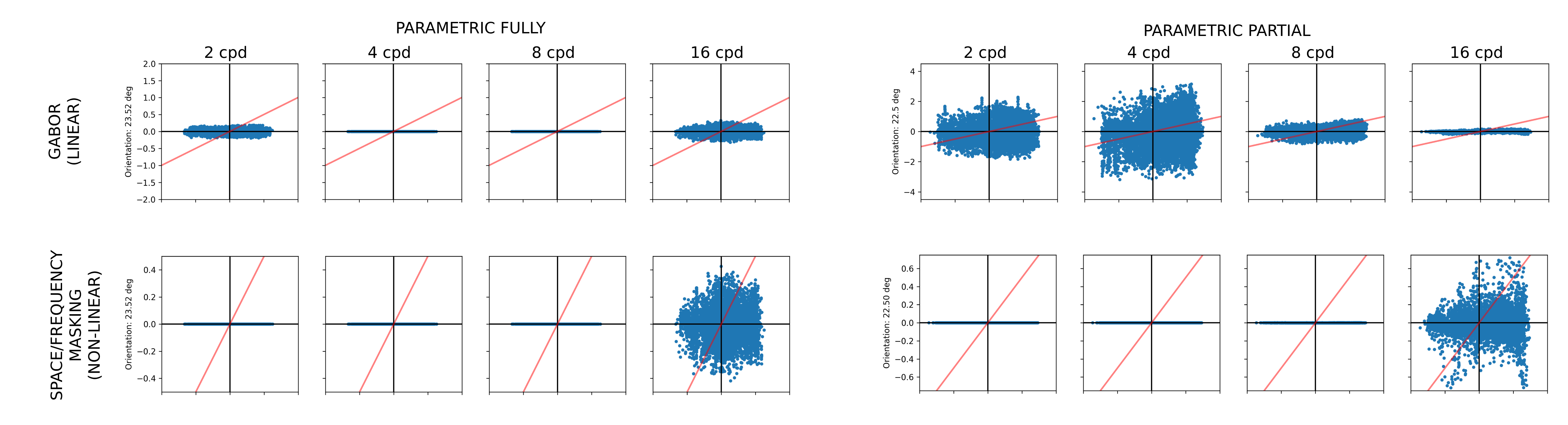}
\caption{\small{\textbf{Input-output scatter plots for the natural-synthetic image}. Linear-V1 (top) and after the divisive normalization and the final scaling (bottom). We only show a subset of all possible input-output responses for the Gabors of 2,~4,~8~and~16 cpd (for a fixed orientation).}}
\vspace{-0.5cm}
    \label{fig:gabor_resp}
\end{figure*}

\begin{figure*}
\centering
\hspace{-1cm}\includegraphics[width=1\textwidth]{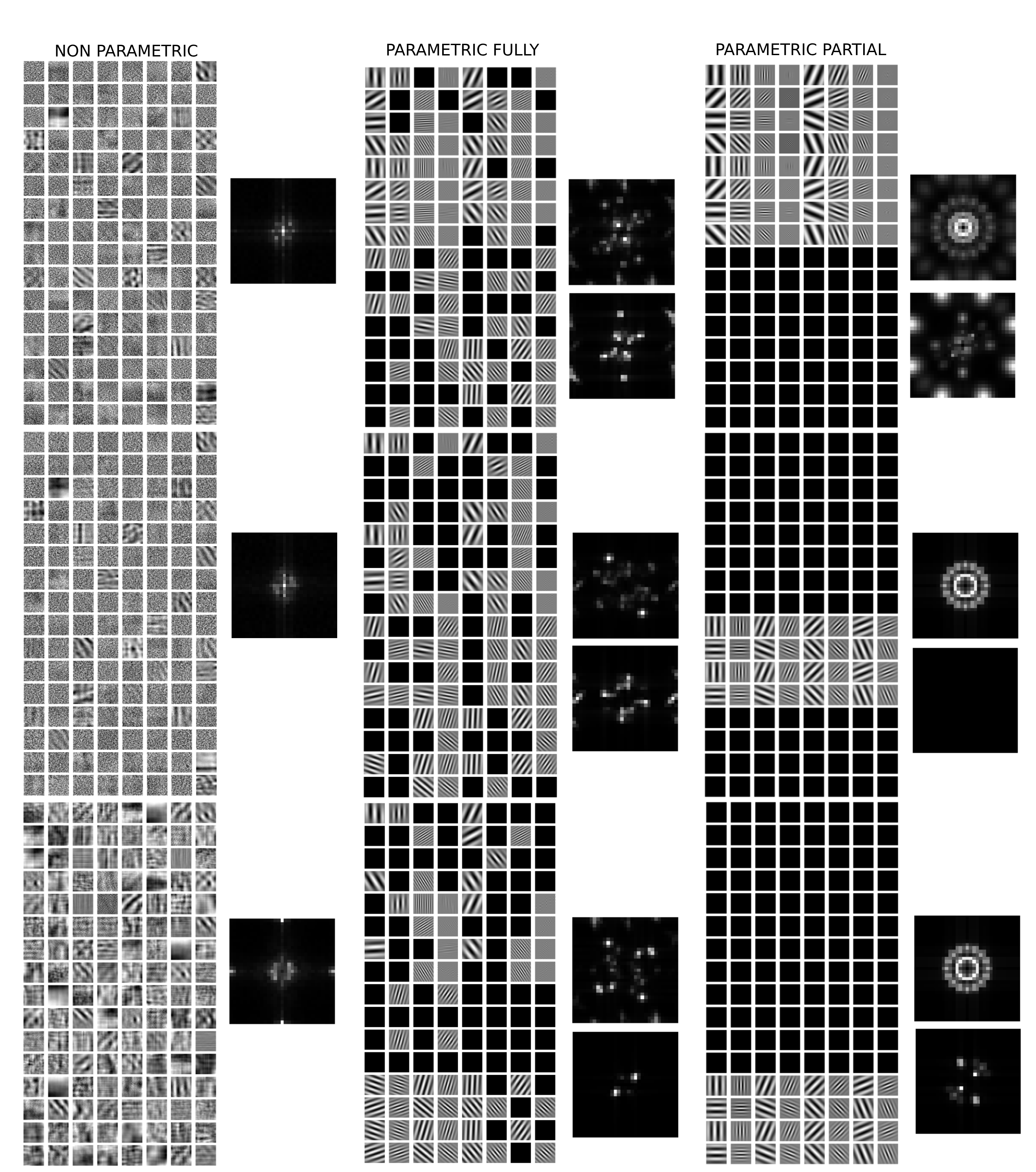}
\caption{\small{\textbf{Model Parameters (or kernels) for the last Gabor-like layer}. Each of the three columns represents the results for each of the three models: non-parametric, parametric bio-fitted, and parametric psycophisically fitted. Each row shows the filters for each of the input channels. In the bio-fitted model the filters act independently to each input channel. The Fourier transform of all the filters combined are given for each model and channel.}}
    \label{fig:D}
\end{figure*}

In Figure \ref{fig:gabor_resp} we observe the input-output scatter plots of the different Gabor filters before and after the application of the last divisive normalization (including the final scaling of Layer~8). Results for the non-parametric model are not shown since they can not be interpreted due to the lack of understanding of the previous layer (see Fig.~\ref{fig:D}). The only behavior biologically plausible is the band-pass trend of the Param-Partial model despite the CSF was not explicitly imposed. %Taking into account that the distribution of samples on the x-axis has the same amplitude regardless of spatial frequency, and that  we have the response of filters of different frequencies on the y-axis,
Filters sensitive to relatively low frequencies, 2~cycles/deg (cpd), amplify the signal in a moderate way, while those of medium frequencies (4~cpd) are the ones that amplify it the most. This is consistent with the fact that the maximum of the contrast sensitivity function is found around a frequency of 4~cpd~\cite{Campbell68,Li22,Malo97a}. When we consider higher frequencies, the amplitude of the responses progressively falls, indicating the bandpass character of the CSF of the achromatic channel. This behavior is not visible in the Param-Fully model shown in the upper left panel.
It is very interesting that, after the application of the scaled divisive normalization, both models reach a totally equivalent gain in these different channels, indicating that this particular configuration is optimal for maximizing correlation with human opinion in the databases used, even if this particular distribution of responses makes no biological sense for Gabor filters~\cite{Malo97a}.

% FIG D
%\iffalse

Figure \ref{fig:D} represents the kernels of the linear layer representing the primary visual cortex of each model, together with the sum of their Fourier transforms. %In the parametric cases, the first 64 elements correspond to the achromatic channel, the next 32 to the red-green channel, and the last 32 to the blue-yellow channel.
For the parametric models we have included the sum of the Fourier transforms weighted by the amplitude of the transformation of the divisive normalization of the last cortical layer. We can see that in the Param-Partial model the receptive fields are perfectly organized and make biological sense~\cite{Hubel59,Hubel61,Blakemore69,Watson90,Simoncelli92,Olshausen96}. 
%In particular, no interaction between the different achromatic and chromatic channels was taken into account. 
In the Param-Fully model, the weights of each channel have changed so that 
%this is no longer maintained and, therefore, 
the biological sense has been lost. This is evident in the sum of the Fourier transforms of these receptive fields: the Param-Partial model has reasonable contrast sensitivity functions~\cite{Campbell68,Mullen85,Li92,Li22} before being weighted by the final scaling that have been optimized, but applying these weights destroys the behavior that was biologically reasonable. In both the Param-Fully and non-parametric models, the sum of the Fourier transforms of the receptive fields makes no sense either before divisive normalization or after.
%\fi

% FIG F
%\iffalse
\begin{figure*}
    \vspace{-0.0cm}
    \centering
    \includegraphics[width=1\textwidth]{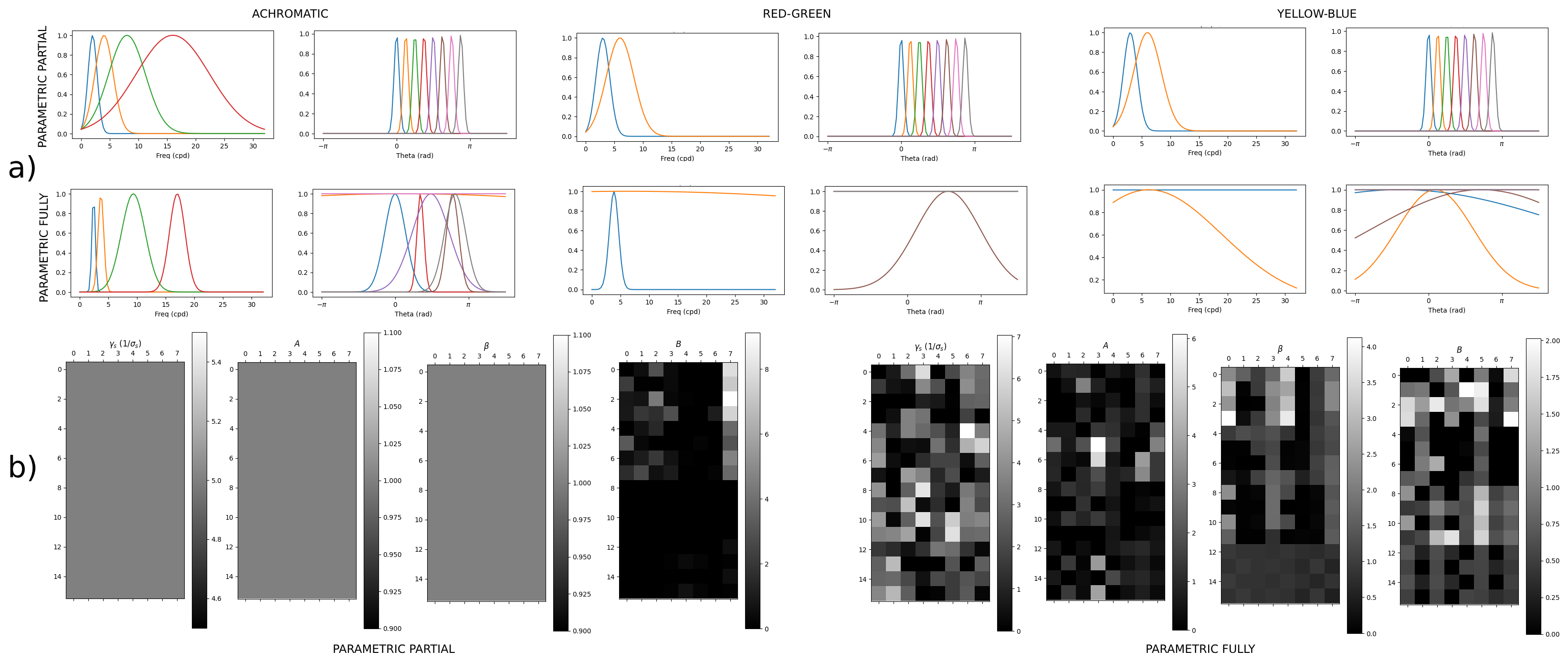}
    \caption{\small{\textbf{Bandwidths (or tuning curves) in frequency and orientation of Gabor filters [a] together with the parameters of cortical divisive normalization [b]}. Each row at the top shows the parameters for each of the parametric models. The first column shows the frequency tuning of the kernels and the second column shows the orientation tuning. Columns in [b] shows the remainder parameters ($\gamma$, $A$, $\beta$, and $B$) for each of the 128 channels.}}
    \vspace{-0.5cm}
    \label{fig:F}
\end{figure*}

Figure \ref{fig:F} shows the parameters for the last layer of the models. For the Param-Partial model the only parameters that are fitted are the weights of the last linear layer. Fitting these weights is needed to adapt the responses to the particular task (or database). Interestingly, for this task the training decides to remove almost completely the color channels as anticipated in Fig.~\ref{fig:B}. While color-channels have less relative importance than the luminance one (as already enforced in the model), this is not the case in the particular task since there are very few color distortions in the database. It is also noticeable that these weights have an effect of enhancing high frequencies (as opposed to the human CSF~\cite{Campbell68,Malo97b}). There are two reasons for these weights to do that: (a) distortions in the database mainly have high frequency components, and (b) the model has two pooling steps which drastically reduce the high frequency components. This is in agreement with the fact that outputs of the Param-Partial and Param-Fully obtain similar output magnitudes after this step (Fig.~\ref{fig:gabor_resp}).
%\fi

\section{Discussion}
\label{sec:discussion}

Results show that using an extremely simple (parametrized) bio-plausible model one can attain state-of-the-art performance in image quality. However, when training all the model parameters, one gets a little bit better results at the task, but its behavior strongly deviates from the bio-plausible solution: 
%%%%%%%%%%%%%%
Figs.~5-10 show that color information is \emph{spread} along the spatial channels and texture information is \emph{spread} along the layers with center-surround neurons and Gabor neurons so that certain high-pass behavior (possibly good for the specific task) is obtained in the end, as shown in Figs.~8 and~9. This effect observed here, which we call \emph{feature spreading}, while irrelevant for the specific task, is a critical problem for the interpretation of the final model. The \emph{feature spreading} may arbitrarily occur along layers if their scale is not controlled, or more generally, if they execute similar functions, not properly constrained by the task, as suggested in Section~III.C.
%%%%%%%%%%%%%%%%%
Figs.~5-10 show that this problem is even more dramatic in non-parametric models (with millions of parameters).   
%%%%%%%%%%%%%%%%%
Our conjecture is that the conventional goal in image quality is not demanding enough to constrain deep models. In fact, both kinds of models (conventional non-parametric and the proposed parametric) achieve correlations bigger than the human upper-bound. This suggests that overfitting is hard to mitigate when the only optimization criterion is correlation.

\begin{figure}[b!]
    \vspace{-0.0cm}
    \centering
    \includegraphics[width=0.5\textwidth]{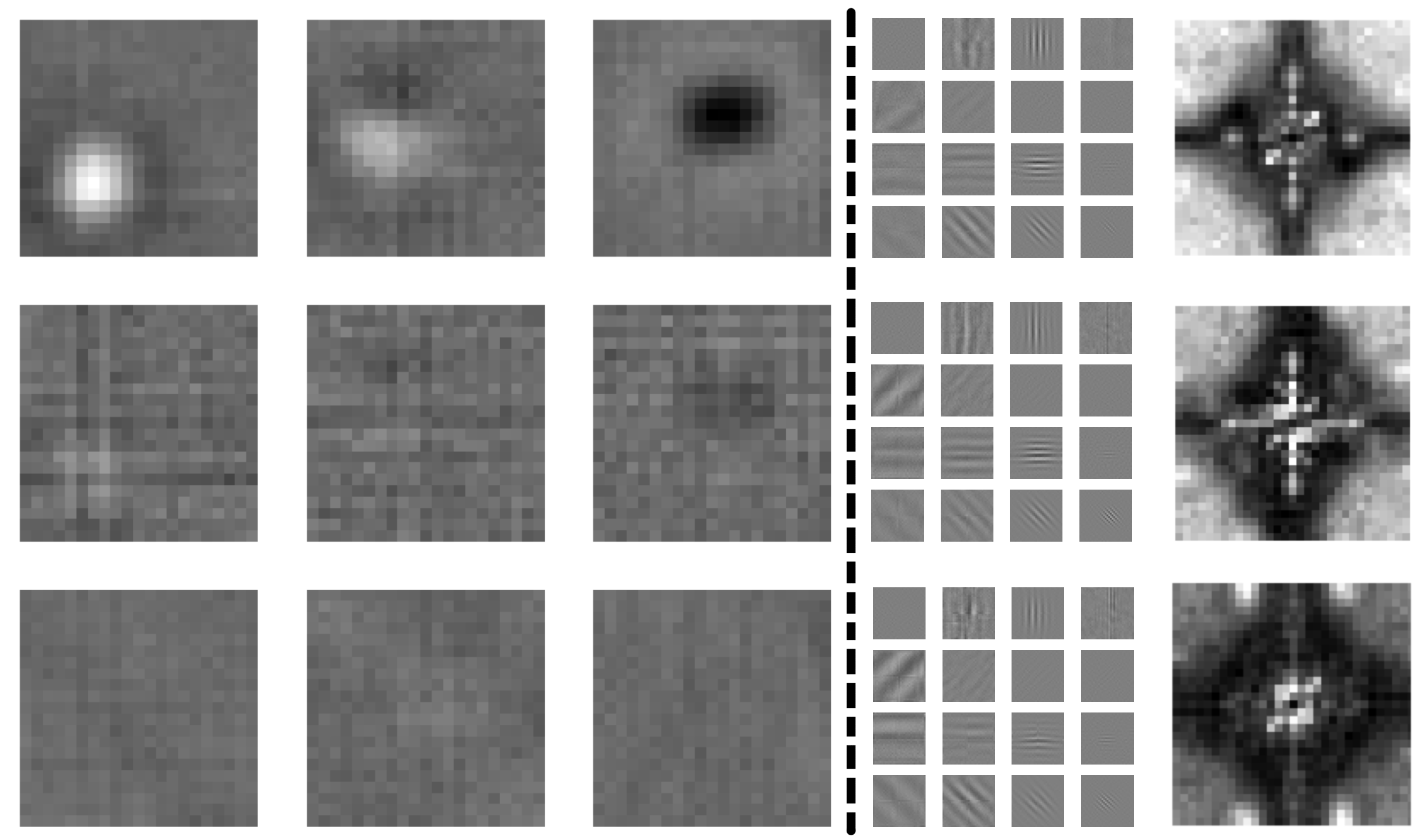}
    \caption{\small{\textbf{Non-parametric model filters when fitted to a more restrictive goal.} Left: Same as the Center-Surround rows in Fig.\ref{fig:C}. Right: Same as in Fig.\ref{fig:D}, and V1-like (right). In both cases the receptive fields show a more bioplausible behavior emerging similar characteristics to LGN and V1.}}
    \label{fig:traca}
\end{figure}

In order to check that intuition, we carry out an additional simulation: we use the non-parametric model to reproduce the joint responses of all neurons in the last layer of a visual system model (the Param-Partial). This more restrictive goal makes the non-parametric model to have more human like behavior: Fig~\ref{fig:traca} shows the receptive fields obtained in the layers of the non-parametric model assumed to be LGN and V1. The first panel shows the emergence of center-surround type receptive fields and low-frequency blobs in the achromatic, red-green, and blue-yellow channels. The obtained receptive fields are consistent with the CSFs~\cite{Campbell68,Mullen85,Li92}. We also see that Gabor-like filters of specific frequencies emerge, consistently with biology~\cite{Hubel59,Hubel61,Blakemore69,Watson90,Simoncelli92,Olshausen96}, and the sum of their Fourier transforms shows that the relevance of the high frequencies is bigger in the achromatic channel, consistently with the human sensitivity for Gabors~\cite{Campbell68,Malo97b}.

This suggests an interesting concept: conventional (non-parametric) deep-nets could also lead to interesting (human-like) models if one could fit their parameters to enough  data.

In any case, we argue that parametrization is the way to go to understand what is going on at each stage. Parametric models suffer less from the \emph{feature-spreading problem}, and allow us to check if certain qualitative behaviors hold. 

\section{Conclusions and Final remarks}
\label{sec:conclusions}

This work proposes an early vision model within the deep learning framework. The model has parametric bioinspired layers, and we proposed a set of suitable bioinspired parameters for the model. Therefore the model is fully interpretable in vision science terms. It obtains state-of-the-art behavior in image quality task (in test) while reducing 4~orders of magnitude the number of parameters regarding the non-parametric version.

We analyze three comparable versions of the proposed architecture in vision science terms: a non-parametric one, and two parametric versions with biologically plausible initializations fitted with different degrees of freedom. The analysis of the models shows that the proposed parametric model with bioinspired layers and parameters holds all the tested human vision requirements. 
However, the non-parametric model can not be interpreted as a vision model. On the other hand, the parametric model with bioinspired initialization deviates from being an interpretable vision model when it is fully trained to maximize correlation. We suggest that this \emph{feature spreading} problem will happen whenever redundant layers are stacked and optimized for goals that are not restrictive enough. 

This parametric exploration not only has implications for practical applications but also opens up new research directions in the field of computational modeling of the human visual system. In particular, we foresee four important avenues for further research. 
(1)  The obvious way is improving the model including new perceptual effects: while the proposed architecture includes multiple facts about the human visual system, we restricted ourselves to maintain the architecture in \cite{Hepburn20} in order to make comparisons fair. However, many modifications are still possible. For instance, the front-end of the model that carries out adaptation to luminance and color could be better: we did not include variable scaling of the photo-receptors according to changes in spectral illumination (i.e. Von-Kries adaptation~\cite{Fairchild13}, or the crispening effects~\cite{Whittle92}). Right now, only takes into account the Weber law. Moreover, it does not take into account subtractive color adaptation in the opponent channels~\cite{Krauskopf92,Gutmann14}. This would imply more general versions of the DN, as shown in the Supplementary Material~F. More importantly, it would require that the critical scaling of the nonlinear layers be input-dependent. Suggestions in this regard have already been given~\cite{Martinez19}, and they could lead to better convergence of the models too. 
(2) A different step forward  is proposing new architectures where the considered layers are more independent. A possibility to get compact architectures with more independent modules is following the spirit in~\cite{Amari92}. 
The problem of dependence between parameters has been pointed out in this kind of retina-V1 models when using time-consuming psychophysics to estimate them~\cite{Malo15}: interaction between parameters of different layers increases the length of the experimental sessions that optimize one layer at a time.
(3) Another new avenue could be connecting bioinspired attention mechanisms (which basically work at the level of V1~\cite{Itti98,Li02}) with the attention mechanisms that are at the core of transformers~\cite{Vaswani17}.
(4) Finally, results in section \ref{sec:discussion} suggest using more restrictive training procedures. As pointed out in~\cite{bowers2023deep}, the rich empirical literature on visual psychophysics should be comprehensively taken into account, rather than just relying on fitting massive datasets (through scalar correlation) that may not represent those critical facts in the models~\cite{Rust06,Martinez19}. 
Low-level~\cite{Malo24charla,Malo22b} and high-level~\cite{Bowers24} datasets have been compiled and could be used to this end.
Recent works already reported that too simple performance measures in reasonable visual tasks may lead to non-human behaving models~\cite{Geirhos20,Geirhos18,Li22,Arash23,GomezVilla20,Pablo_NeuralNets_25}.

\balance

\iffalse
\section*{Acknowledgments}
This work was supported in part by MICIIN/FEDER/UE under Grant PID2020-118071GB-I00 and PDC2021-121522-C21, by Spanish MIU under Grant FPU21/02256 and in part by Generalitat Valenciana under Projects GV/2021/074, CIPROM/2021/056, and by the grant BBVA Foundations of Science program: Maths, Stats, Comp. Sci. and AI (VIS4NN). Some computer resources were provided by Artemisa, funded by the EU ERDF through the Instituto de Física Corpuscular, IFIC (CSIC-UV). 
\fi

\section*{Data and code Availability Statement}
Code to reproduce the results is in the Github repository: %\href{https://github.com/Jorgvt/PerceptualTests}{https://github.com/Jorgvt/paramperceptnet}.
\small{\texttt{https://github.com/Jorgvt/paramperceptnet}}

\appendices

\section{}
\textbf{Supplementary Material} contains details as Sections~A-F. See: \small{\texttt{http://isp.uv.es/docs/ParamPerceptNetSupplem.pdf}}

\bibliographystyle{IEEEtran}
%\bibliography{biblio_clean,actual5}
\bibliography{biblio_cleaned}

%%%%%%%%%%%%%%%%%%%%%%%%%%%%%%%%%%
%%%%%%%%%%%%%%%%%%%%%%%%%%%%%%%%%%
%%%%%%%%%%%%%%%%%%%%%%%%%%%%%%%%%%

\end{document}

\title{Parametric PerceptNet: A bio-inspired deep-net \\ trained for Image Quality Assessment (Supplementary Material)}

\author{Jorge Vila-Tomás\textsuperscript{1}, Pablo Hernández-Cámara\textsuperscript{1}, Valero Laparra\textsuperscript{1}, Jesús Malo\textsuperscript{1} \\
\textit{\textsuperscript{1}Image Processing Lab, Universitat de València, Paterna, 46980, Spain}}

\maketitle

\begin{abstract}
In this supplementary material we provide: 
(A)~an in-depth explanation of the parametric forms chosen to be used in the model described in the main document, 
(B)~specific parametrization options for these functional forms,
(C)~a psychophysically meaningful initialization for each of the parameters,
(D)~the description of the natural-synthetic image used as illustrative stimulus for visualizing the responses of the proposed model,
(E)~visualization of the stimuli that maximize the response of the layers tuned to texture, and 
(F)~details on generalizations of divisive normalization for further improvements of parametric models.

This was separated from the main document due to maximum length requirements.

\end{abstract}

\begin{IEEEkeywords}
Image Quality Metrics, Human Vision, Parametric Neural Networks
\end{IEEEkeywords}

\subsection{\textbf{Parametric layers in deep nets based on Early Vision}}
\label{sec:intro_param}

In this section we describe the layers based on known functional forms that will be used by our model. We will define both the linear (based on convolutions) and the non-linear parts. For the linear parts we will define a functional form for the convolutional kernel, and for the nonlinear parts we will use a bio-inspired layer called divisive normalization (which can also include functional convolutions in its calculations) \cite{Carandini12, Balle16, Malo24}.

%{\bf VAL: esto lo he traido de la intro, hay que darle todavía una vuelta.} 
Recent efforts to include parametrized biological computations into the models by construction are scarce
~\cite{Luan17,Alekseev19,DiCarlo20,Perez20,veerabadran21,Zara21,EVANS202296,chen2023}.
And those approaches are very limited because they mainly rely on constraining, at most, partial aspects of the architecture and optimizing all the other model's parameters blindly~\cite{Luan17,Alekseev19,DiCarlo20,Perez20,veerabadran21,Zara21,EVANS202296,chen2023}.

\subsubsection{Convolutional Filters}

Usually a convolutional filter is defined as a tensor with $D_x \times D_y \times D_i \times D_z$ dimensions, where $D_x$ and $D_y$ are the spatial dimensions, and $D_i$ and $D_z$ are the input and output channels dimensions respectively. Instead of defining each position of the tensor as an isolated parameter (which is the conventional approach), here we are going to define parametrized functions that provide the values for each position of the tensor as a function of a reduced set of generative parameters. 

%The general idea is as follows,
We define a spatial function $F$ in the spatial image dimensions (similar to pixel position) where we have two coordinates: $x \in R^{D_x}$ and $y \in R^{D_y}$. The specific shape of $F$ is determined by some trainable parameters, $\Theta$. Note that, opposite to the classical definition of the convolutional filter in neural networks, the number of parameters does not depend on the filter size. 

For the input channels dimension we repeat the same spatial form, but we scale the values by a parameter $A_{iz} \geq 0$, i.e. $g(x,y,i,z,\Theta_z) = A_{iz} G(x,y,\Theta_z)$, similar to depthwise separable convolution. One could think on finding a different set of parameters $\Theta$ for each channel. However, we found our parametrization more convenient for different reasons: first, the number of parameters is smaller in our implementation. Second, it seems reasonable to combine features of the same spatial form of different channels. And third, this allows us to interpret the meaning of the learned filter more easily.  

Each channel output of the convolutional linear layer is:
\begin{equation}
    X^{n+1}(x,y,z) = \mathcal{F}_{CL}(X^n,\Theta) = g(x,y,i,z,\Theta) * X^n(x,y,i)
\label{eq:convolution}
\end{equation}

Where $X^n$ and $X^{n+1}$ are the layer input and output respectively, and $*$ is the convolution operation in the spatial dimensions.
%Appendix \ref{sec:app_param_formulations} details the mathematical expressions of the functions chosen to parametrize $G$.
Below we define the spatial functions chosen $G$, and how they are parametrized. As a general rule, the functional forms are centered with respect to the filter and normalized so that their energy equals 1.

\paragraph{Gaussian Filter.}
\label{sec:gaussian}

A basic core function to the other functional forms that we use in this work is the classical Gaussian function: 
\begin{equation}
    G(x,y,\Theta) = e^{- \frac{1}{2} \left(\frac{(x-x_0)^2}{\sigma_x^2} + \frac{(y-y_0)^2}{ \sigma_y^2} \right)}
\label{eq:gaussian}
\end{equation}

This filter has two shape parameters, $\sigma_x$ and $\sigma_y$, i.e. $\Theta = \{\sigma_x, \sigma_y\}$, and one scaling factor per input channel, $A_{iz} = diag(A_{z})$. In practice, we optimize $\gamma = 1/\sigma$ for numerical stability.

\paragraph{Difference of Gaussians (DoG) Filter}
\label{sec:dog}

In Retina and Lateral Geniculate Nucleus (LGN) there are neurons that respond to center-surround (or difference of Gaussians) spatial shape \cite{Bonin2005}. It consists of two overlaped Gaussians with a shared center but with different sizes and signs:  

\begin{equation}
    G(x,y,\Theta) =  \frac{1}{\sigma^2} \left[ e^{-\frac{1}{2}  \left(\frac{x^2+y^2}{ \sigma^2} \right)} - \frac{e}{K^2}^{-\frac{1}{2}  \left(\frac{x^2+y^2}{ (K\sigma)^2} \right)} \right]  
\label{eq:dog}
\end{equation}

Center-surround mechanisms do not have orientation selectivity therefore, $\sigma_x= \sigma_y = \sigma$. Our definition (eq. \ref{eq:dog}) contains two parameters, $\sigma$ and $K$, i.e. $\Theta = \{\sigma, K\}$. The scaling factors $A_{iz}$ allows interaction between input channels but we chose to initialize them so that this interactions does not happen, i.e. $A_{iz}=\mathcal{I}$. In practice, we optimize $\gamma = 1/\sigma$ for numerical stability, and restrict $K$ to be bigger than 1 during training to avoid changing the signs between the big and small Gaussians.

\iffalse
\begin{figure}[b!]
    \centering
    \begin{tabular}{cc}
        \includegraphics[width=0.2\textwidth]{Figures/filters/center_surround_095.png} &
        \includegraphics[width=0.2\textwidth]{Figures/filters/center_surround_105.png}
    \end{tabular}
    \caption{Center-surround filters (2D and 3D versions). Left: off-on filter, where the gaussian with smaller $\sigma$ has a negative sign. Right: on-off filter, where the gaussian with smaller $\sigma$ has a positive sign.}
    \label{fig:DoG_filters}
\end{figure}
\fi

\iffalse
\begin{multline}
    G(x,y) = A \left( \frac{1}{2\pi\sigma_1^2} e^{-\left(\frac{(x-x_0)^2+(y-y_0)^2}{2 \sigma_1^2} \right)} - \frac{1}{2\pi\sigma_2^2} e^{-\left(\frac{(x-x_0)^2+(y-y_0)^2}{2 \sigma_2^2} \right)} \right) = \\
    A \left( \frac{1}{2\pi \sigma^2} e^{-\left(\frac{(x-x_0)^2+(y-y_0)^2}{2 \sigma^2} \right)} - \frac{1}{2\pi (K\sigma)^2} e^{-\left(\frac{(x-x_0)^2+(y-y_0)^2}{2 (K\sigma)^2} \right)} \right)
\label{eq:dog}
\end{multline}
\fi

\paragraph{Gabor Filter.}
\label{sec:gabor}

Primary visual cortex (V1) simple cells transform information coming from LGN so that they are orientation selective \cite{Hubel61,Hubel59}. These cells have been modeled classically using a Gabor filter. A Gabor filter is a composition of a Gaussian filter multiplied by a sinusoidal grid:

\begin{equation}
    G(x,y,\Theta) = e^{- \frac{1}{2} \left(\frac{\tilde{x}^2}{\sigma_x^2} + \frac{\tilde{y}^2}{ \sigma_y^2} \right)} \mathrm{cos}(2\pi f (x \: \mathrm{cos}(\theta_f) + y \:\mathrm{sin}(\theta_f))) 
\label{eq:gabor}
\end{equation}

Note that the Gaussian envelope is allowed to be oriented. In order to do so, coordinates are rotated as $[\tilde{x},\tilde{y}] = R(\theta_{\textrm{env}})[x,y]$, where $R(\theta_{\textrm{env}})$ is a rotation matrix with angle $\theta_{\textrm{env}}$. This is equivalent to a symmetric covariance matrix.
%, however it would be more complicated to interpret. 
This way we have two parameters for the Gaussian shape, ${\sigma_x,\sigma_y}$, and one parameter for the Gaussian orientation, $\theta_{\textrm{env}}$. The sinusoidal grid has a frequency parameter, $f$, oriented an angle, $\theta_f$. Note that we allow the Gaussian and the sinusoid to have different orientations. In summary, the whole filter has five shape parameters, i.e. $\Theta = \{\sigma_x,\sigma_y, \theta_{\textrm{env}},f,\theta_f\}$. As in the \emph{DoG}, the scaling values $A_{iz}$ are chosen so that there is no interaction between channels but could be optimized to do so. As in the other layers, we optimize $\gamma = 1/\sigma$ for numerical stability. 
This function allows different degrees of restriction in the parametrization which are discussed below in Section~B.

\subsubsection{Bio-Inspired activation function, Divisive Normalization.}
\label{sec:dn}

Activation functions are usually selected to be simple, fast to compute, dimension-wise and differentiable non-linear functions. While being differentiable and non-linear are mandatory features, being simple, fast to compute and dimension-wise are just convenient for implementation purposes. There are other non-linear activation functions much more aligned with biology. Our selected candidate is the divisive normalization proposed as a canonical operation in the brain \cite{Carandini12}. Therefore, the output for each channel of the nonlinear layer in our model is defined as: 

\begin{equation}
    X^{n+1}(z) = \mathcal{F}_{NL}(X^{n},\Theta_i) = B\frac{X^n}{\left( \beta + G(\Theta) * (X^n)^\alpha \right)^{\varepsilon}}
\label{eq:dn}
\end{equation}

Where $G$ is a convolutional filter that is applied to the input $X^n$ in order to have an estimation of the magnitude of the neighbors. The parameter $\beta$ is a bias parameter that avoids division by zero and controls how non-linear the operation is (among other things). The coefficient $B$ allows the control of the output scaling. We fixed the values of the exponents, to be the same as in \cite{hepburn_perceptnet_2020}, $\alpha = 2$ and $\varepsilon=0.5$. The parameters for each output channel in this layer will be the $B$, $\beta$, and the parameters of the filter $G$. In general $G$ will be a Gaussian filter (eq. \ref{eq:gaussian}) which may depend on the parameters of the previous layer. 

One advantage of using parametric functions is that we have direct access to the meaning of the outputs. For instance, we have direct access to the frequency, $f$, and the orientation $\theta_f$ of each Gabor filter. This can be used to design the following layer taking these parameters as input. This is what we do in the last layer of our model, where the divisive normalization kernel in the denominator depends not only on the spatial position but also on the other channels' frequencies and orientations. The denominator's idea is to have an estimation of the neighbors' magnitude, where these neighbors can be spatial, but can be also frequential, orientation, or chromatic neighbors. We use a simple Gaussian function where the distance in frequency and orientation is also taken into account (see eq. \ref{eq:gaussian_DN}).
As in the previous case, the different degrees of restriction allowed in the parametrization are discussed below.

\subsection{\textbf{Parametrization options}}
\label{sec:model}

{\bf Options for the parametrization of the Gabor layer:} The \emph{Gabor} filter allows different degrees of restriction in the parametrization which involves different number of parameters:
\begin{itemize}
    \item \textbf{All parameters free}: Generate $C_{in} \times C_{out}$ Gabor filters with different parameters for each of them. $\# Params = C_{in} \times C_{out} \times 5$.
    \item \textbf{Repeat for same inputs}: Generate $C_{out}$ Gabor filters with different parameters for each of them and repeat them for every input channel, i.e. If we have 3 input channels and we want to output only 1 feature, we will only generate 1 Gabor and repeat it 3 times to make the convolution. $\# Params = C_{out} \times 5$.
    \item \textbf{Depthwise separable}: A variation of the previous one where the Gabor filter is repeated across the input channels but each one of them is weighted by a learned parameter $A$, multiplying the amount of parameters by a factor $C_{in}$. This is our selection which is consistent with the depthwise separable idea. $\# Params = C_{out} \times (5+C_{in})$.
    \item \textbf{Fixed set of $f$ and $\theta$}: Define a set of $f$ and $\theta$ and generate the Gabors corresponding to all their possible combinations ($N_{f} \times N_{\theta}$ filters). Each of them will have an independent Gaussian envelope. As before, each Gabor is repeated $C_{in}$ times to perform the convolution. $\# Params = N_{f} + N_{\theta} + (N_{f} + N_{\theta}) \times 3$ (params of the gaussian) $= (N_{f} + N_{\theta}) \times 4$.
    \item \textbf{Fixed set of $f$ and $\theta$ (Restricted envelope)}: All the previous parametrizations allow the Gaussian envelopes to have their long side perpendicular to the sinusoid, but this is not what we find in humans. To enforce a more human-like structure we can parametrize $\sigma_y = 2\sigma_x$, reducing even more the amount of parameters. $\# Params = N_{f} + N_{\theta} + (N_{f} + N_{\theta}) \times 2$ (params of the gaussian) $= (N_{f} + N_{\theta}) \times 3$.
\end{itemize}

\vspace{0.5cm}
{\bf Options for the parametrization of the DN layer:} We can employ different forms of the canonical Divisive Normalization (which corresponds to choosing different forms form the operator $H$). Keep in mind that the output of the Divisive Normalization has the same number of channels as the input ($z = i$).
\begin{itemize}
    \item \textbf{Original}: Dense interaction between pixels and features. $\# Params = k_{x} \times k_{y} \times D_i \times D_o = k_{x} \times k_{y} \times D_i^{2}$, where $k_x$ and $k_y$ are the width and height of the convolutional filter respectively. This is the form to be found in the original PerceptNet model \cite{hepburn_perceptnet_2020}.
    \item \textbf{Spatial Gaussian}: Gaussian interaction between pixels and dense interaction between features.
    \begin{itemize}
        \item \textbf{With repetition}: Defining $C_{in}$ Gaussians and repeating over the $C_{in}$ features. $\# Params = C_{in} \times 1$. (Remember that a Gaussian has only 1 parameter).
        \item \textbf{Without repetition}: $\# Params = C_{in} \times C_{in} = C_{in}^{2}$.
    \end{itemize}
    \item \textbf{Spatial-Features Gaussian}: Gaussian interaction between pixels and features. $\# Params = C_{in} \times C_{in} = C_{in}^{2}$.
    \item \textbf{Chromatic-Spatio-Temp-Orient Gaussian}: Gaussian interaction between pixels and features taking into account the features' chromaticities, frequencies, and orientations. Requires that the chromaticities, frequencies and orientations of the input features are known (something that we get if using parametric Gabor layers and chromatic decomposition layers as well). The expression is shown in \ref{eq:gaussian_DN}. $\# Params = C_{in} \times N_{f} \times N_{\theta}$.
\end{itemize}

\begin{equation}
    G(x,y,\Theta) = H e^{- \frac{1}{2} \left(\frac{x^2+y^2}{\sigma_s^2} + \frac{(f-f_0)^2}{\sigma_f^2} + \frac{(\theta-\theta_0)^2}{ \sigma_\theta^2} \right)}
\label{eq:gaussian_DN}
\end{equation}

\subsection{\textbf{Psychophysically meaningful initialization}}
\label{app:init}

Following the literature exposed in Section II of the main text (see Fig.~1 in the main text) we set the parameters of each layer as described in table \ref{tab:init_params}. 

The spatial and chromatic calibration of the images is important because visibility of distortions depends on the observation distance (or subtended visual angle), and the values of the parameters depend on the range of the input.
In our case, the model assumes a sampling frequency of $f_s=64$~(cpd), which means that the visual angle subtended by a $256\times256$ image is 2 degrees. The model works with images of any size assuming the corresponding visual angle for the mentioned sampling frequency. RGB digital values are assumed to be normalized in the [0,1] range.

The first layer applies a Divisive Normalization (DN) point-wise nonlinearity to the inputs, initialized to be consistent with the Weber law~\cite{Stiles00}.

The second linear layer transforms the normalized RGB values in every pixel into opponent colors: achromatic (A), red-green (T), and yellow-blue (D), following the standard Jameson \& Hurvich color space~\cite{Jameson57,Vila23}.

The third layer applies a point-wise DN nonlinearity to the ATD value in every pixel, according to the nonlinearities pointed out by Krauskopf \& Gegenfurtner~\cite{Gegen92}.

The fourth layer convolves the ATD images using center-surround neurons using Differences of Gaussians (DoGs) with the parameters shown in the table. These are consistent with the achromatic, red-green and yellow-blue CSF filters~\cite{Campbell68,Mullen85}, which are assumed to be originated by the LGN cells~\cite{Li92,Li22}.

\iffalse
Central column in Figure \ref{fig:C} shows the effect of this choices in the parameters for the first 5 layers, the same for the layer 6 in Fig.~\ref{fig:D}, and layer 7 in Fig.~\ref{fig:F}. 

The effect of each layer over a natural image is shown in figures \ref{fig:A} and \ref{fig:B}. Fig.~\ref{fig:C} first row shows that the first layer enhances the sensitivity for low luminances and reduces it for high luminances. Fig.~\ref{fig:C} second row shows the parameters of the color space change, it is a matrix multiplication that changes to opponent colors: one achromatic, and two chromatics (red-green, and yellow-blue). Fig.~\ref{fig:C} third row shows how the normalization in the third layer accounts for a contractive function in the achromatic channel while keeping the chromatic channels unchanged. Fig.~\ref{fig:C} forth row shows the center-surround filters, the values for these filters have been selected in order to perform a band-pass filter according with the achromatic and chromatic CSF. Values for the parameters in layer 5 perform a spatial normalization which takes into account close neighbours. These relations follow a spatial Gaussian (Fig.~\ref{fig:C} row 5), and have a saturation effect which depends on the neighbours energy (Fig.~\ref{fig:C} row 6). 
\fi

The fifth nonlinear layer equalizes the outputs of the center-surround neurons using a DN transform in the spatial domain~\cite{Ahumada92,Watson02,Malo15,Martinez18}. 

Following~\cite{Simoncelli92}, the Gabors for 6th layer have been chosen to have 8 linearly spaced orientations, 4 scales separated in octaves, and independent filters for the achromatic and each chromatic channel. Since the visual system is more sensitive to achromatic than chromatic stimuli, we use half of the filters for the achromatic channel and half for both chromatic channels, as shown in Fig.~9 (right) of the main text. 

Finally, following~\cite{Schwartz01,Laparra10,Malo10,Martinez19,Malo24} the DN of the 7th layer were chosen to normalize the input taking into account the activity of close neighbors in frequency, orientation, and space. 

The parameter $B$ of the final 8th layer is the one that determines the importance to each output channel. It is initially assumed to be the identity for every channel and then this scale is fitted to optimize the task. 

\begin{table}[t!]
\caption{\small{\textbf{Psychophysical initialization of the parametric model}. The sampling frequency is assumed to be 
$f_s=64$~(cpd), which means that the visual angle subtended by a $256\times256$ image is 2 degrees.
RGB digital values are assumed to be normalized in the [0,1] range. The symbol $\ldots$ means this initial value is applied to every element, and then independently optimized thereafter.}}

\label{tab:init_params}
\begin{center}
\begin{tabular}{|c|c|c|c|}

\hline
\multicolumn{4}{|c|}{\bf Layer 1 (DN - "RGB Normalization")} \\
\hline
 & R & G & B \\
\hline
$\beta$ & \multicolumn{3}{c|}{0.1} \\
\hline
$H$ & \multicolumn{3}{c|}{0.5} \\
\hline\hline
\multicolumn{4}{|c|}{\bf Layer 2 (Conv - "Jameson \& Hurvich Opponent Colors")} \\
\hline
\multicolumn{4}{|c|}{
\[
\begin{bmatrix}
A \\
T \\
D
\end{bmatrix} =
\begin{bmatrix}
0.24 & 0.71 & 0.10 \\
0.18 & -0.39 & 0.19 \\
0.09 & 0.25 & -0.38
\end{bmatrix}
\cdot
\begin{bmatrix}
R \\
G \\
B
\end{bmatrix}
\]}\\
\hline\hline
\multicolumn{4}{|c|}{\bf Layer 3 (DN - "Color Normalization")} \\
\hline
 & A & T & D \\
\hline
$\beta$ & 1 & 1 & 1 \\
$H$ & 1/3 & 1/3 & 1/3 \\
\hline\hline
\multicolumn{4}{|c|}{\bf Layer 4 (Conv "DoG")} \\
\hline
 & A & T & D \\
\hline
A & 1, 0, 0 & 0, 1, 0 & 0, 0, 1 \\
K & 1.1, 1.1, 1.1 & 5, 5, 5 & 5, 5, 5 \\
% $\mathrm{log}\sigma$ & -1.9, -1.9, -1.9 & -1.76, -1.76, -1,76 & -1.76, -1.76, -1,76  \\
$\mathrm{log}\sigma$ & -1.9... & -1.76... & -1.76... \\
\hline\hline
\multicolumn{4}{|c|}{\bf Layer 5 (DN - "LGN Normalization")} \\
\hline
 & A & T & D \\
\hline
A & 1 & 1 & 1 \\
$\beta$ & 0.1 & 0.1 & 0.1 \\
$\gamma$ & 25 & 25 & 25 \\
\hline\hline
\multicolumn{4}{|c|}{\bf Layer 6 (Conv - "V1 Gabor"} \\
\hline
 & A & T & D \\
\hline
freq. (cpd) & 2, \, 4, \, 8,\, 16 & 3,\, 6 & 3,\, 6 \\
$\gamma_{fx}$ & 1.87,  3.48,  6.50, 12.13 & 2.69, 5.02 & 2.69, 5.02  \\
$\gamma_{fy}$ & 1.49, 2.79, 5.20, 9.70 & 2.15, 4.01 & 2.15, 4.01  \\
\hline
$\theta_{f} (rad)$ & \multicolumn{3}{c|}{0, $\pi$/8, 2$\pi$/8, 3$\pi$/8, 4$\pi$/8, 5$\pi$/8, 6$\pi$/8, 7$\pi$/8} \\
\hline
$\theta_{\textrm{env}}$ (rad)& \multicolumn{3}{c|}{0, $\pi$/8, 2$\pi$/8, 3$\pi$/8, 4$\pi$/8, 5$\pi$/8, 6$\pi$/8, 7$\pi$/8} \\
\hline\hline
\multicolumn{4}{|c|}{\bf Layer 7 (DN - "V1 Normalization") } \\
\hline
A & \multicolumn{3}{c|}{1...} \\
$\gamma_{s}$ & \multicolumn{3}{c|}{5...} \\
\hline
 & A & T & D  \\
\hline
$\gamma_{f}$ & 1.25, 0.63, 0.31, 0.16 & 0.83 , 0.42 & 0.83 , 0.42  \\
\hline
$\sigma_{o}$ & \multicolumn{3}{c|}{0.11$\pi$,0.11$\pi$,0.11$\pi$,0.11$\pi$,0.11$\pi$,0.11$\pi$,0.11$\pi$,0.11$\pi$} \\
\hline
H & \multicolumn{3}{c|}{$\mathbf{1}$ (Identity Matrix)} \\
\hline
\hline
\multicolumn{4}{|c|}{\bf Layer 8 (Linear scale of the final DN) } \\
\hline
$B$ & \multicolumn{3}{c|}{1...} \\
\hline
\end{tabular}
\end{center}
\end{table}

\subsection{\textbf{Properties and generation of the illustrative image used in the experiments}}

\begin{figure}[t!]
    \centering
    \includegraphics[width=0.32\textwidth,height=0.32\textwidth]
    %,height = 0.2\textwidth]
    {Figures/input_image.png}
\caption{\textbf{\small{Illustrative natural-synthetic image for the experiments}}. The natural part of the image contains objects in opponent chromatic directions and a shadow that modifies average luminance and the contrast of the scene. The synthetic part of the image is also affected by the shadow and consists of achromatic sinusoids of 4 and 8 cpd of different orientations (the image subtends 4 degrees).  See Section~D for details on its generation.}
    \label{fig:A}
\end{figure}

Figure~\ref{fig:A} shows the image used in the visualization experiments of the main text.
This image has been chosen to illustrate a range of known psychophysical phenomena.
First, the natural part of the image (the standard Parrots image~\cite{KodakDatabase}) was chosen because it has objects whose colors (basically) are in the known opponent red-green and yellow-blue directions~\cite{Jameson57,Vila23} and can be used to check if there are layers tuned to these relevant visual facts. Then,
the periodic patterns were chosen to have  
4 and 8 cpd respectively~\footnote{In this example the sampling frequency has been chosen assuming that the image is observed at a distance so that it subtends 4 degrees of visual angle.}, and they are oriented at 0 and 45 degrees. These textures (known to induce optimal stimulation in certain regions of the V1 brain~\cite{Blakemore69,Daugman,Watson97}) were chosen to check if the responses of the different layers are specifically tuned to these kind of patterns.
Then, a controlled shadow has been added to the picture. This shadow is generated through controlled variations of the luminance and the contrast of the chromatic channels across the image.
First we reduced the luminance in the left part of the image in order to check the the known Weber-law, which enhances the response in the low luminance regions and moderates the response in the high luminance regions~\cite{Weber,Stiles00,Fairchild13}. Finally, the contrast of the left part has also been reduced to check if the models fulfill the (known) fact that visual system increases its sensitivity (or boosts the responses) in the low-contrast regions~\cite{Watson97,Carandini12,Martinez18,Martinez19}. 
While the luminance has been changed by taking into account the tristimulus-to-digital-values correction and simply reducing the mean in the achromatic channel of a psychophysically-meaningful opponent color space~\cite{Jameson57} (keeping the mean of the chromatic channels the same), the modification of the contrast is more delicate. 
This is because contrast and colorfulness perception is not independent of the luminance~\cite{Mante05,Fairchild13}. Therefore, contrast has been modified by using an RMSE definition of contrast based on the ratio between the standard deviation and the mean of an image which is applicable to natural images and reduces to the classical Michelson contrast for sinusoids~\cite{Peli90}. 
This modification was not only made in the achromatic channel (as mentioned above for the luminance), but also applied in the chromatic opponent channels not to introduce non-natural variations of colorfulness in the image. 
The Matlab code to generate this image is available here\footnote{\texttt{loretes.m} which uses the Matlab toolboxes Colorlab~\cite{Colorlab02} and Vistalab~\cite{vistalab}.}.

\subsection{\textbf{Feature visualization of texture-sensitive layers}}

\begin{figure*}
    \centering
    \includegraphics[width=0.92\textwidth]{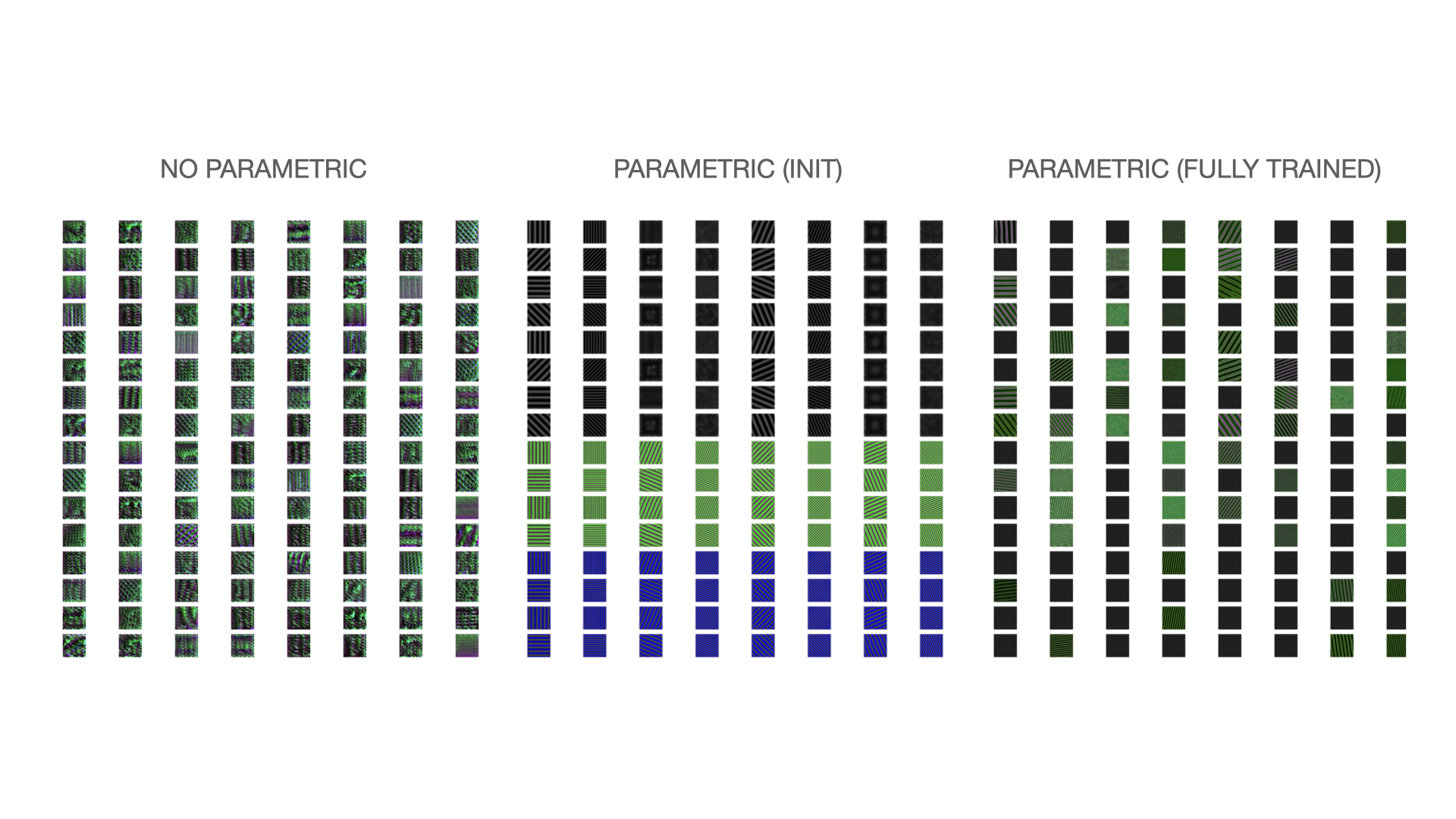}
    \caption{\small{\textbf{Images that maximize the response of each channel 
    of layer 6}.  Each of the three columns represents the results for each of the three models: the non-parametric (where kernels have no specific shape -see Fig.~9 in the main text-), and in the parametric models where kernels are Gabor functions with specific frequency and orientation (and also a specific chromatic content in the param-partial model).}}
    \label{fig:E}
\end{figure*}

Enforcing Gabor filters at a particular layer is not equivalent to measuring Gabor-like receptive fields in a particular brain region because this brain measurement takes into account all the processing steps done up to that point.
Therefore, one could argue that the textures that optimally stimulate the Gabor kernels may differ from the kernels in frequency, orientation or chromatic content.  
In order to address this question, we obtain the input image that maximizes the output of every channel independently. The results are shown in Fig.~\ref{fig:E} and confirm that our proposed initialization (in the parametric-partial model) is tuned to sinusoids with the expected frequency and orientation. Moreover, from the chromatic perspective,
the optimal images in the Param-Partial model are  achromatic for the channels that are supposed to be achromatic, and saturate the digital channels green and blue (respectively) for the channels that are supposed to be tuned to red-green direction (T) and the yellow-blue direction (D).
Note the 64 achromatic filters and 64 chromatic ones (32 green-red and 32 blue-yellow). In contrast, in the other models the achromatic and the chromatic information is spread along the different channels. Moreover, optimal textures are biased to the green-red after optimization in both cases (probably induced by a bias in the training set). 

\subsection{\textbf{Possible improvements in the color/brigthness front-end}}

Compatibility with the base-line non-parametric model~\cite{Hepburn20} restricted ourselves to the use of simple expressions for the Divisive Normalization, thus neglecting well known facts about adaptation to luminance and color.
Fig.~\ref{fig:further} shows obvious improvements: the model presented here does not include variable scaling of the photo-receptors according to changes in spectral illumination (i.e. Von-Kries adaptation~\cite{Fairchild13}, or the crispening effects pointed out by Whittle~\cite{Whittle92}), but only takes into account the Weber law nonlinearity. Moreover, it does not take into account subtractive color adaptation in the opponent channels~\cite{Krauskopf92,Gutmann14}. This would imply more general versions of the DN, as shown in the figure (bottom row). 
These expressions include terms that compute the average of the image through an all-ones matrix \textbf{1} to get an input-dependent DN that implies shifts in the curves that happen in humans when changing the average luminance and the color of the illuminant. More importantly, it would require that the critical scaling of the nonlinear layers, the \emph{A's} in the presented notation, be input-dependent. Suggestions in this regard have already been given~\cite{Martinez19}, and as suggested by Martinez et al. in their hand-crafted exploration, they could lead to better convergence of the models too.

In a more general vein, parametric alternatives to the Divisive Normalization have been proposed for this adaptive front end, which not only reproduce crispening, as in~\cite{Bertalmio20}, but also reproduce interesting texture effects~\cite{Bertalmio24}. 

\begin{figure*}[h!]
    \centering
\includegraphics[width=0.88\textwidth]{Figures/Fig_Further.png}
\caption{\small{\textbf{Better luminance/color front end.} Top: simplified front-end proposed in this work for fair comparison  with~\cite{Hepburn20} (figures with experimental data reproduced from~\cite{Laparra12}). These nonlinearities can be fitted with the DN expressions used here. 
Bottom: possible improvements in the DN expressions to include crispening for different background luminance (experiemntal data reviewed in~\cite{Bertalmio20}) and subtractive adaptation in the chromatic channels (experimental data reviewed in~\cite{Laparra12}). Following the behavior reported in~\cite{Gegen92}, greenish, blueish and yellowish illuminations would give shifts in the responses indicated by the curves of the corresponding colors, thus requiring subtractions.}}
    \label{fig:further}
\end{figure*}

\iffalse
$$
y = A \frac{x}{\beta + Hx}
$$

$$
y = A\frac{x}{\beta + H |x|}
$$

$$
y = A \frac{x}{\beta + (I+\alpha\mathbf{1})Hx}
$$

$$
y = A\frac{(I - \alpha\mathbf{1})x}{\beta + (I - \alpha\mathbf{1}) H |x|}
$$
\fi

\subsection{\textbf{Model Parameters for the last Gabor-like layer.}}

\begin{figure*}
\centering
\hspace{-1cm}\includegraphics[width=1\textwidth]{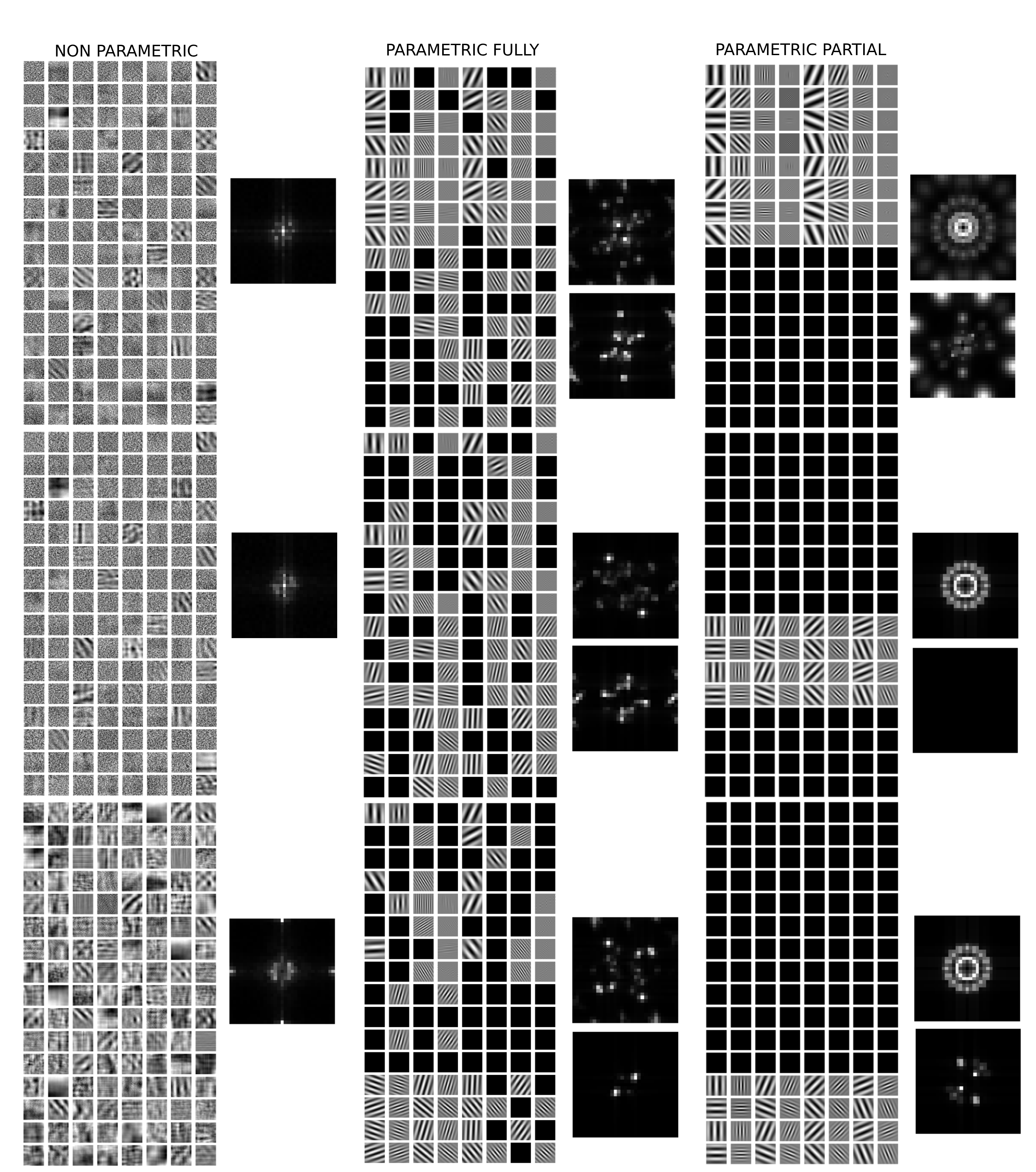}
\caption{\small{\textbf{Model Parameters (or kernels) for the last Gabor-like layer}. Each of the three columns represents the results for each of the three models: non-parametric, parametric bio-fitted, and parametric psycophisically fitted. Each row shows the filters for each of the input channels. In the bio-fitted model the filters act independently to each input channel. The Fourier transform of all the filters combined are given for each model and channel.}}
    \label{fig:D}
\end{figure*}

Figure \ref{fig:D} represents the kernels of the linear layer representing the primary visual cortex of each model, together with the sum of their Fourier transforms. %In the parametric cases, the first 64 elements correspond to the achromatic channel, the next 32 to the red-green channel, and the last 32 to the blue-yellow channel.
For the parametric models we have included the sum of the Fourier transforms weighted by the amplitude of the transformation of the divisive normalization of the last cortical layer. We can see that in the Param-Partial model the receptive fields are perfectly organized and make biological sense~\cite{Hubel59,Hubel61,Blakemore69,Watson90,Simoncelli92,Olshausen96}. 
%In particular, no interaction between the different achromatic and chromatic channels was taken into account. 
In the Param-Fully model, the weights of each channel have changed so that 
%this is no longer maintained and, therefore, 
the biological sense has been lost. This is evident in the sum of the Fourier transforms of these receptive fields: the Param-Partial model has reasonable contrast sensitivity functions~\cite{Campbell68,Mullen85,Li92,Li22} before being weighted by the final scaling that have been optimized, but applying these weights destroys the behavior that was biologically reasonable. In both the Param-Fully and non-parametric models, the sum of the Fourier transforms of the receptive fields makes no sense either before divisive normalization or after.

\bibliographystyle{IEEEtran}
\balance
\bibliography{biblio_clean,actual5,main}

%%% Please see the test.bib file for some examples of references